\documentclass[10pt,twocolumn,letterpaper]{article}

\usepackage{configs/cvpr}  

\definecolor{cvprblue}{rgb}{0.21,0.49,0.74}
\usepackage[pagebackref,breaklinks,colorlinks,allcolors=cvprblue]{hyperref}

\def\paperID{xxx} 
\def\confName{xxx}
\def\confYear{xxx}

\title{TGSFormer: Scalable Temporal Gaussian Splatting for Embodied Semantic Scene Completion}
 
\author{
Rui Qian$^1$$^{^*}$, Haozhi Cao$^1$$^{^*}$, Tianchen Deng$^2$$^{^*}$, Tianxin Hu$^1$, Weixiang Guo$^1$, Shenghai Yuan$^1$, Lihua Xie$^1$$^{\dagger}$ \\
$^1$Nanyang Technological University, $^2$Shanghai Jiaotong University 
}

\begin{document}

\twocolumn[{
\maketitle 
\vspace{-10pt}
\begin{minipage}{\textwidth}
\centering
\includegraphics[width=0.99\linewidth]{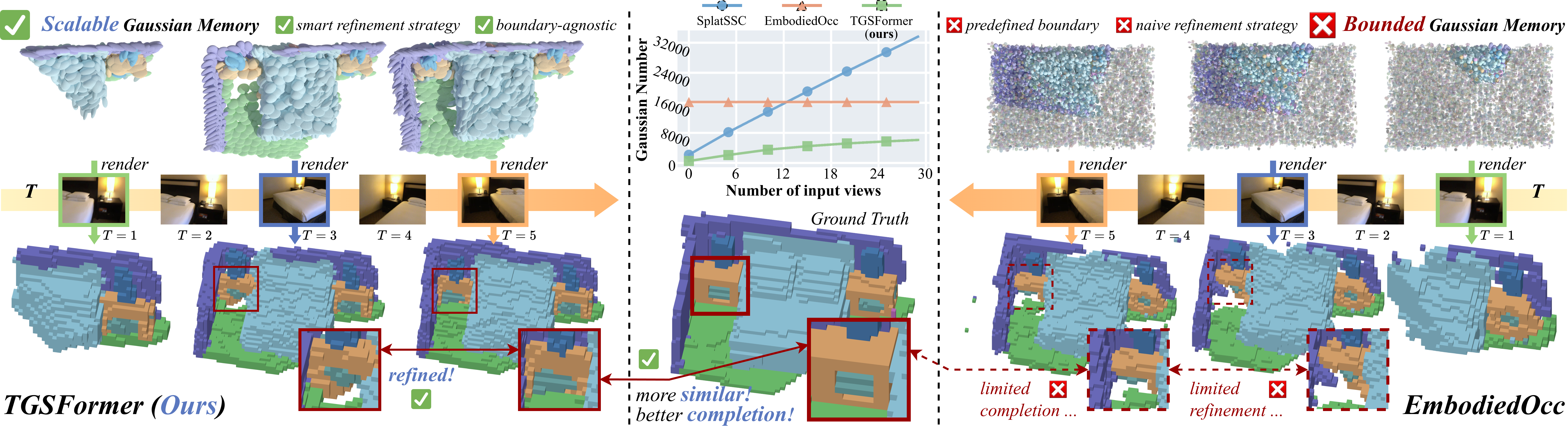}
\captionsetup{type=figure, skip=6pt} 
\captionof{figure}{ 
\textbf{Overview of embodied scene exploration and refinement.}
Our TGSFormer consistently expands its understanding of the environment as new views are observed and progressively refines previously seen regions, producing a complete and coherent 3D scene.
} 
\label{fig:intro}
\end{minipage}
\vspace{6pt}
}]

\begin{abstract}
 
Embodied 3D Semantic Scene Completion (SSC) infers dense geometry and semantics from continuous egocentric observations. Most existing Gaussian-based methods rely on random initialization of many primitives within predefined spatial bounds, resulting in redundancy and poor scalability to unbounded scenes. 
Recent depth-guided approach alleviates this issue but remains local, suffering from latency and memory overhead as scale increases.
To overcome these challenges, we propose \textbf{TGSFormer}, a scalable \textbf{\underline{T}}emporal \textbf{\underline{G}}aussian \textbf{\underline{S}}platting framework for embodied SSC. It maintains a persistent Gaussian memory for temporal prediction, without relying on image coherence or frame caches.
For temporal fusion, a Dual Temporal Encoder jointly processes current and historical Gaussian features through confidence-aware cross-attention. 
Subsequently, a Confidence-aware Voxel Fusion module merges overlapping primitives into voxel-aligned representations, regulating density and maintaining compactness.
Extensive experiments demonstrate that TGSFormer achieves state-of-the-art results on both local and embodied SSC benchmarks, offering superior accuracy and scalability with significantly fewer primitives while maintaining consistent long-term scene integrity. 
The code will be released upon acceptance. 

\end{abstract}
 
\section{Introduction}
\label{sec:intro}

Understanding and reconstructing 3D environments~\cite{huang2023tri, wei2023surroundocc, tian2023occ3d, li2023fb, wang2024panoocc, cao2024pasco, wang2024sfpnet} from continuous egocentric observations is fundamental for embodied perception~\cite{deng2024plgslam, deng2024compact, deng2025mcnslammultiagentcollaborativeneural, deng2025mne} and long-horizon autonomy~\cite{deng2023long, deng2025vpgs}.
Embodied 3D Semantic Scene Completion (SSC)~\cite{wu2024embodiedocc, zhang2025roboocc, wang2025embodiedoccplusplus} aims to infer dense geometry and semantics of a scene from a stream of first-person views, 
enabling agents to perceive, plan, and interact within unexplored environments in the downstream tasks. 
Compared with the conventional single-scan monocular reconstruction~\cite{yu2024monocular, cao2024monoscene}, embodied SSC requires both geometric expressiveness and temporal stability, while maintaining scalability and efficiency over the exploration process. 

Rapid progress has been achieved in both local and embodied SSC during the past few years, most of which can be categorized into two types: (i) dense voxel-based approaches and (ii) sparse object-centric approaches. 
The former~\cite{cao2024monoscene, zhang2023occformer, wei2023surroundocc} considers dense 3D voxels as the elemental representation, while it suffers from computational inefficiency. 
In contrast, object-centric~\cite{huang2024gaussianformer,huang2025gaussianformer2, zhao2025gaussianformer3d} paradigms leverage sparse elements, such as Gaussian primitives, to enhance efficiency and maintain competitive performance. 

When it comes to embodied scenarios, several challenges are posed for existing Gaussian-based methods.
To enable long-term embodied predictions, a crucial procedure is to initialize and maintain a prediction memory of the explored areas, allowing the output of each timestep to interact with its historical counterparts. 
Most embodied SSC methods~\cite{wu2024embodiedocc,wang2025embodiedoccplusplus,zhang2025roboocc, xian2025discene} initialize dense Gaussian primitives within a predefined bounded volume to cover the exploration area, which leads to \textit{redundancy} and \textit{inefficiency}. 
This strategy also becomes \textit{infeasible} when boundary priors are unknown, limiting scalability in real-world exploration. 
A recent alternative~\cite{ruiqian2025splatssc} introduces a depth-guided initialization scheme for local SSC, but it lacks a mechanism for maintaining long-term prediction, causing \textit{noise accumulation} and \textit{memory overhead} as observations increase.
Moreover, while the spatial–temporal approaches~\cite{yan2025stgs,leng2025stocc} perform well in short-term temporal fusion, they rely on frame-to-frame coherence, making their predictions \textit{fragile when key frames are missing or inconsistent}. 
 
To this end, we propose \textbf{TGSFormer}, a scalable temporal Gaussian splatting framework for embodied scene completion. 
For each incoming image, an effective memory paradigm should preserve reliable predictions while keeping the representation compact during exploration. Accordingly, TGSFormer employs a persistent Gaussian memory that accumulates scene representations over time and updates them through feature association instead of frame caching.
The Dual Temporal Encoder (DTE) fuses current and historical Gaussian features via confidence-aware cross-attention, ensuring stable temporal reasoning and reliable information flow.
Afterwards, the Confidence-aware Voxel Fusion (CAVF) module merges primitives in differentiable voxel space, regulating density while maintaining bounded memory and structural compactness.
In addition, multi-stage supervision aligns intermediate Gaussian features, ensuring consistent representation across layers. The model is trained with monocular pretraining, followed by embodied fine-tuning, supporting stable temporal adaptation without additional overhead. 
This feed-forward architecture enables scalable, memory-efficient and temporally consistent scene completion across embodied environments, avoiding reliance on continuous temporal images. 
 
Our main contributions are summarized as follows:
\begin{itemize}
    \item We present a temporal feed-forward Gaussian-based SSC framework with a persistent and compact 3D memory, enabling borderless global scene completion without relying on image coherence, and bridging local depth-guided SSC with large-scale embodied perception.
    \item We introduce a dual-temporal encoder with confidence-aware cross-attention, which aligns current and historical Gaussian primitives for reliable temporal fusion.
    \item We develop a training-free differentiable voxel Gaussian fusion module that performs confidence-weighted merging to ensure compact and bounded memory. 
    \item We propose multi-stage supervision to align features, enhancing cross-level geometric–semantic consistency and stabilizing embodied reconstruction. 
\end{itemize}


\section{Related Works}

\subsection{Sparse 3D Scene Representation.}
3D occupancy scene representation~\cite{roldao2020lmscnet, tian2023occ3d,yu2023flashocc, hou2024fastocc, tang2024sparseocc, cao2024pasco, shi2025oneocc} aims to infer spatially complete and semantically meaningful 3D structures from partial visual observations. Recent efforts have increasingly emphasized sparse and efficient modeling to address the scalability bottleneck of dense voxel representations. 
VoxFormer~\cite{li2023voxformer} introduced a sparse-to-dense transformer pipeline that generates 3D proposals from geometry priors before hierarchical voxel refinement. Subsequent works~\cite{mei2024camera, zhu2024CGFormer, jiang2024symphonize} further advanced this paradigm through context-aware attention, unified encoder-decoder design, and instance-level reasoning, establishing sparse voxel modeling as a new foundation for vision-based 3D scene representation.

A recent trend moves beyond voxel discretization toward object-centric representations. 
GaussianFormer~\cite{huang2024gaussianformer} pioneered this paradigm by representing 3D scenes with continuous Gaussian primitives, and rendering them into volumetric semantics via Gaussian-to-voxel splatting, effectively capturing scene sparsity while remaining differentiable. 
Follow-up works~\cite{huang2025gaussianformer2, zhao2025gaussianformer3d} introduced probabilistic formulations and LiDAR-guided initialization, while EmbodiedOcc~\cite{wu2024embodiedocc} and its extensions~\cite{zhang2025roboocc, wang2025embodiedoccplusplus, xian2025discene} further adapted this paradigm to online perception through confidence-based and geometry-aware refinements. 
This evolution from sparse voxels to sparse Gaussians forms the conceptual basis of our TGSFormer framework.

\subsection{Spatial-Temporal 3D Scene Modeling.}
Temporal cues are crucial for robust vision-centric occupancy perception, yet dense voxel fusion incurs substantial memory and computation. Early BEV-based temporal models align and reuse historical features via recurrent self-attention or feature concatenation~\cite{hang2021bevdet, li2022bevformer, liu2023sparsebev}, and recent occupancy pipelines extend these ideas from 2D BEV to 3D volumes, which limits the number of fused frames due to cubic costs~\cite{li2023fb, tian2023occ3d, wang2024panoocc}. 
To improve efficiency and long-term consistency, ST-Occ~\cite{leng2025stocc} introduces a unified scene-centered spatiotemporal memory with uncertainty-aware attention, achieving lower temporal inconsistency with reduced overhead. 
Along a complementary line in the sparse Gaussian paradigm, ST-GS~\cite{yan2025stgs} enhances spatial interaction and temporal coherence for Gaussian primitives through dual-mode aggregation and geometry-aware fusion, demonstrating that temporal modeling is equally beneficial beyond voxel grids. 

Different from these methods, TGSFormer builds upon the frame-agnostic embodied paradigm~\cite{wu2024embodiedocc} and jointly handles initialization, temporal fusion, and memory bounding within a unified Gaussian representation, enabling stable long-term updates and scalable global scene completion.

\begin{figure*}[!htbp]
\centering 
\vspace{-10pt}
\includegraphics[width=0.98\textwidth]{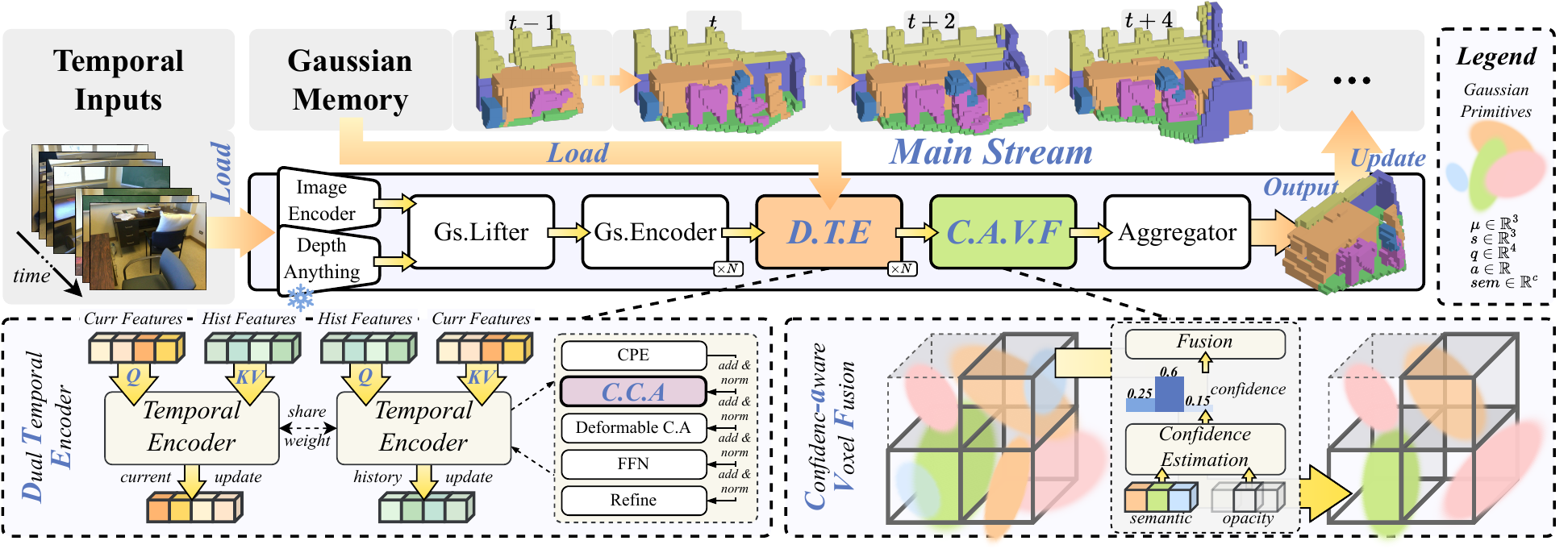} %
\vspace{-10pt}
\caption{
\textbf{An overview of our proposed TGSFormer architecture}. Our framework first employs parallel image and depth encoders to extract appearance features and geometry priors. These are passed to a Gaussian Lifter (Gs.Lifter) and a Gaussian Encoder (Gs.Encoder) to generate the current set of Gaussian primitives and embeddings.
These primitives are then fed into our \textit{Dual Temporal Encoder} (DTE). The DTE loads historical features queried from the global Gaussian Memory and processes both data streams using two weight-sharing Temporal Encoders.
The fused representations are passed to our \textit{Confidence-aware Voxel Fusion (CAVF)} module, which estimates per-primitive semantic and opacity uncertainty, then performs a confidence-weighted fusion to merge primitives and control density.
Finally, an aggregator splats the merged Gaussians into the semantic voxel grid. These primitives are then used to update the global Gaussian Cache. 
}
\label{fig:pipeline}
\vspace{-8pt}
\end{figure*} 

\section{Methodology}
\label{sec:formatting}
The architecture of our approach is illustrated in Fig.~\ref{fig:pipeline}, which can be divided into two phases: (1) monocular local prediction and (2) Gaussian memory maintenance. In this section, we first present the preliminary and problem setup of embodied SSC (Sec.~\ref{sec:problem_setup}). Subsequently, we briefly introduce our coarse local predictions procedures (Sec.~\ref{sec:loc_pred}) and detail the following Gaussian memory maintenance as the core design of our TGSFormer (Sec.~\ref{sec:gaussian_mt}).
 
\subsection{Preliminary and Problem Setup}

\label{sec:problem_setup}
\noindent \textbf{Scene Representation Preliminary.} 
Our primary goal is to estimate the volumetric occupancy and semantics of a 3D voxel grid $\mathcal{V}$ given only the 2D image input. Our method models the scene with $N$ 3D Gaussian primitives $\mathcal{G} = \{G_i\}_{i=1}^N$. Specifically, each Gaussian primitive $G_i$ is parameterized by its geometric and semantic properties. These include a mean $\boldsymbol{\mu}_i \in \mathbb{R}^3$, a scale $\mathbf{s}_i \in \mathbb{R}^3$, a rotation $\mathbf{q}_i \in \mathbb{R}^4$, an opacity $\mathbf{a}_i \in [0, 1]$, and semantic logits $\mathbf{c}_i \in \mathbb{R}^{C-1}$. The scale and rotation define the anisotropic covariance matrix $\mathbf{\Sigma}_i$:
\begin{equation}
    \mathbf{\Sigma}_i = \mathbf{R}_i \mathbf{S}_i \mathbf{S}_i^T \mathbf{R}_i^T, \mathbf{S}_i = \mathrm{diag}(\mathbf{s}_i), \mathbf{R}_i = \mathrm{q2r}(\mathbf{q}_i),\label{eq:covariance}
\end{equation} 
where $\mathrm{q2r}(\cdot)$ converts quaternions to rotation matrices.
 
We render the sparse set $\mathcal{G}$ into a dense grid $\mathcal{V}$ using a decoupled Gaussian-to-voxel splatting scheme~\cite{ruiqian2025splatssc}. For each voxel center $\mathbf{x}$, the rendering process computes the empty probabilities $\alpha(\mathbf{x})$ and semantic probabilities $e^l(\mathbf{x})$ by aggregating contributions from a neighborhood $\mathcal{N}(\mathbf{x})$:
\allowdisplaybreaks 
\begin{align}
    \alpha(\mathbf{x}) &= 1 - \prod_{i \in \mathcal{N}(\mathbf{x})} \left(1 - \alpha(\mathbf{x};G_i) \cdot \mathbf{a}_i\right), \\
    e^l(\mathbf{x}) &= \frac{\sum_{i \in \mathcal{N}(\mathbf{x})} p(\mathbf{x}|G_i) \cdot \tilde{\mathbf{c}}_i^l}{\sum_{j \in \mathcal{N}(\mathbf{x})} p(\mathbf{x}|G_j)}, \\
    p(\mathbf{x}|G_i) &= \frac{1}{(2\pi)^{3/2} |\mathbf{\Sigma}_i|^{1/2}} \alpha(\mathbf{x}; G_i).
\end{align}
Here, $\alpha(\mathbf{x}; G_i) = \exp(-\frac{1}{2}(\mathbf{x}-\boldsymbol{\mu}_i)^\top \mathbf{\Sigma}_i^{-1} (\mathbf{x}-\boldsymbol{\mu}_i))$ is the un-normalized Gaussian kernel and $\tilde{\mathbf{c}}_i$ are the softmax-normalized probabilities derived from logits $\mathbf{c}_i$. The final voxel probabilities are then defined as:
\begin{equation}
    \hat{\mathcal{V}}_x^l = \alpha(\mathbf{x}) \cdot e^l(\mathbf{x}), \quad \hat{\mathcal{V}}_x^{\mathrm{empty}} = 1 - \alpha(\mathbf{x}). \label{eq:final_voxel}
\end{equation}
 
\noindent \textbf{Problem Setup.}
\label{sec:embodied_prediction}
Unlike conventional indoor SSC methods~\cite{song2017sscnet, wang2019forknet, wang2023semantic, yu2024monocular} which primarily optimize the monocular predictions, our proposed TGSFormer focuses on embodied SSC~\cite{wu2024embodiedocc}. In this more challenging scenario, agents are expected to gather stream-like visual input and update their understanding of the environment online. Formally, we consider an agent receiving a continuous stream of observations $\mathcal{X} = \{x_1, x_2, \dots, x_t, \dots\}$, each of which contains the current RGB image and camera pose, denoted as $x_t = \{\mathcal{I}_\mathrm{rgb}^t, \mathcal{P}^t\}$. The agent is trained to maintain a persistent global Gaussian memory $\mathbb{M}_t$ that represents the semantic understanding of the explored area.

\subsection{Coarse Local Prediction} 
\label{sec:loc_pred}
Given the input image at timestep $t$, TGSFormer performs monocular SSC prediction largely following the existing depth-guided approach SplatSSC~\cite{ruiqian2025splatssc}. Specifically, the multi-scale features are extracted by the image encoder~\cite{tan2019efficientnet, lin2017feature}. Depth features and geometry prior are additionally supplied by a finetuned \textit{DepthAnythingV2}~\cite{yang2024depth} model. These features are fed to our modified Gaussian Lifter to generate initial Gaussian primitives $\tilde{\mathcal{G}}_t$ and embeddings $\tilde{\mathcal{Q}}_t$. $\{\tilde{\mathcal{G}}_t, \tilde{\mathcal{Q}}_t\}$ is then refined by a series of Gaussian Encoder (GSE) blocks to produce local coarse representation $\{\mathcal{G}_t, \mathcal{Q}_t\}$. 

\noindent \textbf{Modified Gaussian Lifter.}
\label{sec:gaussian_lifter} 
We experimentally observe that a direct lifting strategy is highly effective when paired with robust depth priors. With strong depth cues from \textit{DepthAnythingV2}~\cite{yang2024depth}, the modified lifting approach can avoid complex components, such as geometry-aware modules~\cite{wu2024embodiedocc} or multi-modal fusion steps~\cite{ruiqian2025splatssc}, and in practice delivers significantly stronger SSC performance in local prediction (as shown in Tab.~\ref{tab:ablation_gaussian_initialization}). 
 
To elaborate, similar to SplatSSC, our process applies the uniform sampling on $\mathcal{F}_d$ and $\mathcal{I}_\mathrm{d}$ to obtain a sampled depth map $\mathcal{I}_\mathrm{d}^\mathrm{s}$ and corresponding features $\mathcal{F}_\mathrm{d}^\mathrm{s}$, where $\mathcal{I}_\mathrm{d}^\mathrm{s}$ is then reprojected using camera intrinsics $K \in \mathbb{R}^{3\times 3}$ to define the initial Gaussian means $\boldsymbol{\mu}_o$, with other primitive attributes randomly initialized. The features $\mathcal{F}_\mathrm{d}^\mathrm{s}$ are used directly as the initial Gaussian embeddings $\tilde{\mathcal{Q}}_t$.

\subsection{Gaussian Memory Maintenance}
\label{sec:gaussian_mt}
While the depth-guided local prediction yields strong local SSC, it alone cannot enforce temporal consistency or keep the number of primitives bounded as exploration proceeds. We therefore introduce a principled Gaussian memory mechanism that preserves and refines historical representations over time. The core design of TGSFormer is the efficient and effective interaction between the current local primitives and this historical Gaussian memory. To enable unbounded exploration without prohibitive overhead, TGSFormer maintains an accumulated Gaussian memory and updates the historical state $\mathbb{M}_{t-1}$ as: 
\begin{align} 
    \mathbb{M}_{1} &= \{\mathcal{G}_{1}, \mathcal{Q}_{1}\}, \\ 
    \mathbb{M}_t &= \mathcal{M}_\mathrm{TGSFormer}(x_t, \mathbb{M}_{t-1}),
\end{align}
when $t=1$, the local predictions serve as the initial historical Gaussian memory. Given the local predictions $\{\mathcal{G}_t, \mathcal{Q}_t\}$, we first retrieve the historical primitives from $\mathbb{M}_{t-1}$ that fall within the field-of-view of the current frame $x_t$, denoted $\{\hat{\mathcal{G}}, \hat{\mathcal{Q}}\}$.
The Dual Temporal Encoder (DTE) performs temporal fusion over these two sets and produces refined historical primitives 
$\{\hat{\mathcal{G}}', \hat{\mathcal{Q}}'\}$ together with refined local primitives $\{\mathcal{G}_t', \mathcal{Q}'_t\}$. 
These outputs are then merged through the Confidence-aware Voxel Fusion (CAVF) module, whose results update the global Gaussian memory.

It is also worth noticing that when our maintenance applies to monocular SSC, where the query operation degenerates to $\{\hat{\mathcal{G}}, \hat{\mathcal{Q}}\} = \{\mathcal{G}_t, \mathcal{Q}_t\}$, transiting to a self-refinement paradigm for improved local predictions. 

\begin{figure}[!t]  
\centering
\vspace{-4mm}
\includegraphics[width=0.95\columnwidth]{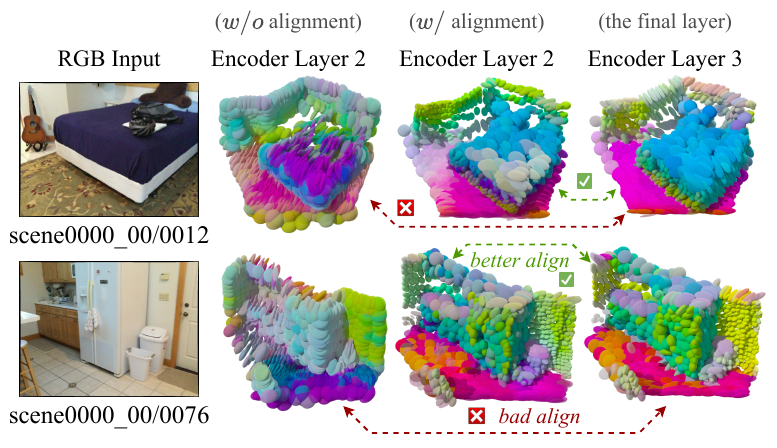}
\vspace{-10pt} 
\caption{\textbf{Feature alignment visualization with Principal Component Analysis (PCA).} PCA projections of Gaussian features show that our multi-stage objective not only aligns intermediate representations toward the final encoder space, but also makes their distributions more isotropic and semantically organized. 
}
\label{fig:pca_analysis} 
\vspace{-10pt}
\end{figure} 

\noindent \textbf{Dual Temporal Encoder.}
\label{sec:temporal_encoder} 
To effectively process temporal information and enable continuous perception in complex scenes, we propose DTE, which can be flexibly applied to both embodied and single-frame prediction. For embodied scenarios, DTE facilitates dual-stream cross-interactions between local predictions and historical Gaussian memory by considering their corresponding prediction confidence. Specifically, DTE employs two parallel weight-sharing temporal encoders, one of which queries the historical primitives $\{\hat{\mathcal{G}}, \hat{\mathcal{Q}}\}$ with the current primitives $\{\mathcal{G}_t, \mathcal{Q}_t\}$, and vice versa for the other. Each Temporal Encoder augments the standard GSE structure with a \textit{Confidence-aware Cross Attention} (CCA) module. In the dual-stream embodied mode, CCA performs confidence-aware cross-attention to produce updated features $\mathcal{Q}_t^c$ and $\hat{\mathcal{Q}}^c$: 
\begin{align}
    \mathcal{Q}_t^c &= \texttt{CCA}(\mathcal{Q}_t, \hat{\mathcal{Q}}, C_{t}, \hat{C}), \label{eq:cca_local}\\
    \hat{\mathcal{Q}}^c &= \texttt{CCA}(\hat{\mathcal{Q}}, \mathcal{Q}_t, \hat{C}, C_{t}),
\end{align}
where $C_{t}$ and $\hat{C}$ are the confidence of $\mathcal{G}_t$ and $\hat{\mathcal{G}}$ which we elaborate in the following sections. 
When operating in a monocular mode, our DTE performs self-interaction using only the single-frame prediction instead, leading to a one-stream CCA module of self-attention as:
\begin{align} 
    \mathcal{Q}_t^c &= \texttt{CCA}(\mathcal{Q}_{t}, \mathcal{Q}_{t}, C_{t}, C_{t}).
\end{align}
Subsequently, the updated features from both streams are passed through an FFN~\cite{ashish2017transformer} and a refinement module~\cite{huang2024gaussianformer}, yielding the refined current primitives $\{\mathcal{G}_t', \mathcal{Q}'_t\}$ and historical primitives $\{\hat{\mathcal{G}}', \hat{\mathcal{Q}}'\}$.

\begin{figure*}[t]
\centering
\includegraphics[width=0.95\textwidth]{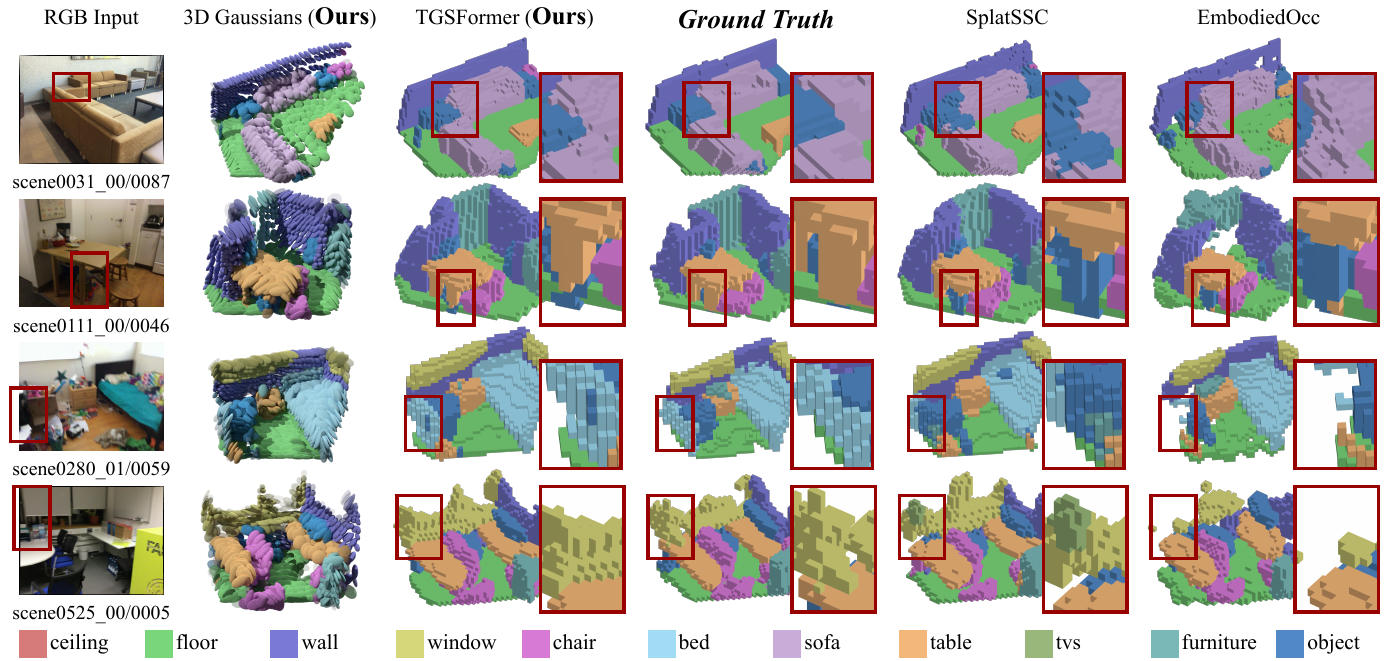} %
\vspace{-6pt}
\caption{\textbf{Qualitative comparison of monocular prediction results on the Occ-ScanNet and Occ-ScanNet-mini dataset.} TGSFormer reconstructs more complete geometry and captures semantics with higher clarity than existing approaches.}
\vspace{-10pt}
\label{fig:vis_base}
\end{figure*}

\noindent \textbf{Confidence Estimation.}
To achieve a more reliable update on both locally predicted primitives and historical memory, we propose to estimate the primitive confidence as an indication of their prediction reliability. Specifically, our confidence estimation modulates information flow through a per-primitive confidence score $C_i \in [0, 1]$ by jointly assessing the semantic uncertainty and geometric stability. 
The semantic uncertainty is quantified using the Shannon entropy $H(\tilde{\mathbf{c}}_i)$, formulated as:
\begin{equation}
    H(\tilde{\mathbf{c}}_i) = -\sum_{k=1}^{C-1} \tilde{\mathbf{c}}_i^k \log(\tilde{\mathbf{c}}_i^k).
\end{equation}
This entropy is converted into a semantic confidence score $C_{\mathrm{sem}}$ via a power transform, with higher entropy mapping to a lower confidence. The final confidence $C_i$ is then computed by multiplying the opacity $\mathbf{a}_i$, which denotes the primitive-wise geometric certainty, formulated as: 
\begin{align}
    C_i = \underbrace{(1 - \min(H(\tilde{\mathbf{c}}_i) / H_{\mathrm{max}}, 1))^p}_{C_\mathrm{sem}} \cdot \mathbf{a}_i, 
\end{align}
where $H_{\mathrm{max}}$ is a maximum entropy hyperparameter and $p$ controls the transform's sharpness.

\noindent
\textbf{Confidence-aware Cross Attention.}
Taking Eq.~\ref{eq:cca_local} as an example, CCA first embeds the input primitive features as:
\begin{equation}
    \mathrm{Q} = \mathcal{Q}_t W_q,\quad \mathrm{K} = \hat{\mathcal{Q}} W_k,\quad \mathrm{V} = \hat{\mathcal{Q}} W_v, 
\end{equation}
where $W_q$, $W_k$, and $W_v$ are trainable weight matrices. Inspired by a recent work~\cite{qiu2025gatedattn}, our confidence modulation is applied at two positions, as shown in Fig.~\ref{fig:modulation-vis}: (1) the historical value $\mathrm{V}$ is modulated by $\hat{C}$ after its linear projection, forming $\mathrm{V}' = \mathrm{V} \odot \hat{C}$, and (2) the aggregated attention output is modulated by $C_t$ before the final output projection $W_o$. Therefore, the CCA for the current feature update branch is formulated as:
\begin{multline}
    \texttt{CCA}(\mathcal{Q}_t, \hat{\mathcal{Q}}, C_{t}, \hat{C}) = \\
    (\texttt{Concat}(\texttt{MHA}(\mathrm{Q}, \mathrm{K}, \mathrm{V}')) \odot C_t)W_o,
\end{multline}
where $\texttt{MHA}(\cdot)$ denotes the standard multi-head attention from FlashAttention-2~\cite{dao2023flashattention2} and $\texttt{Concat}(\cdot)$ is the head concatenate operation. This dual modulation strategy ensures information from high-confidence primitives is trusted and propagated. Information from low-confidence or uncertain primitives is suppressed during temporal fusion.

\begin{figure*}[t]
\centering 
\includegraphics[width=0.98\textwidth]{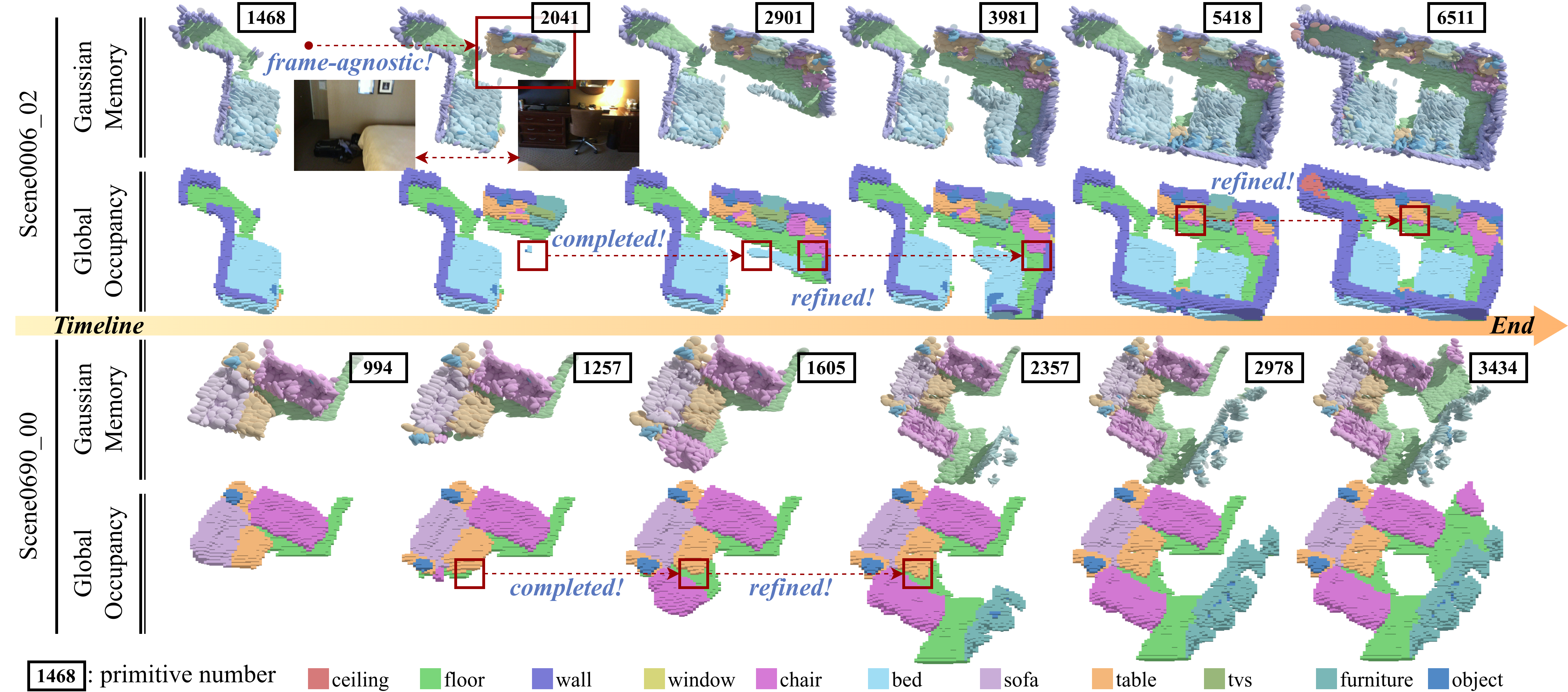} %
\vspace{-6pt}
\caption{\textbf{Global prediction visualization on the EmbodiedOcc-ScanNet-mini dataset.} Our TGSFormer framework not only produces high-quality monocular completion but also consistently refines and completes the observed scene through temporal updates.}
\vspace{-10pt}
\label{fig:vis_glob} 
\end{figure*}

\noindent \textbf{Confidence-aware Voxel Fusion.}
\label{sec:Gaussian_merging}
To achieve robust fusion between $\{\mathcal{G}_t', \mathcal{Q}'_t\}$ and $\{\hat{\mathcal{G}}', \hat{\mathcal{Q}}'\}$, while avoiding the exponential growth of Gaussian primitives in embodied scenes, we introduce the CAVF module upon previous works~\cite{taolu2024scaffoldgs, jiang2025anysplat}. Specifically, CAVF merges all primitives that fall within the same 3D voxel into a single new primitive. In practice, the local predicted and historical primitives are mapped to their corresponding voxel indices $s$ based on their 3D means $\boldsymbol{\mu}_i$ through voxelization, which serves only as a grouping step. The merging weight is then derived from a per-voxel softmax operation applied to the primitive's confidence score $C_i$, computed as:
\begin{equation} \label{eq:cavf_weight}
w_{i \to s} = \frac{\exp(C_i / T)}{\sum_{j: V_j=s} \exp(C_j / T)},
\end{equation}
where $V_j=s$ denotes the set of all primitives $j$ assigned to voxel $s$, and $T$ is the predefined temperature parameter. Subsequently, all attributes (e.g. $\boldsymbol{\mu}, \mathbf{s}, \mathbf{q}, \mathbf{c}$) and features of the new fused primitives $\{\mathcal{G}_s, \mathcal{Q}_s\}$ are computed via a confidence-weighted summation of all included primitives:
\begin{equation} \label{eq:cavf_fuse}
\mathcal{G}_s = \sum_{i: V_i=s} w_{i \to s} \mathcal{G}_i, \quad \mathcal{Q}_s = \sum_{i: V_i=s} w_{i \to s} \mathcal{Q}_i.
\end{equation}
This process significantly reduces the total number of primitives. The resulting set $\{\mathcal{G}_s, \mathcal{Q}_s\}$ is used for the final aggregation and to update the global Gaussian memory.

\subsection{Training Objective}
\label{sec:training_objective}

Direct end-to-end training for embodied prediction often results in unstable convergence, as the model has to learn both single-frame perception and temporal fusion simultaneously. 
We therefore adopt a two-stage strategy: the first stage trains the network on monocular SSC to build a strong perceptual prior, 
and the second stage fine-tunes it for embodied prediction to achieve stable temporal adaptation.

\noindent \textbf{Stage 1: Monocular Pretraining.}
The first stage aims to establish a frame-agnostic perceptual prior. 
Training frames are randomly sampled across different scenes to remove temporal correlation. The model is trained to infer per-frame semantic occupancy from a single RGB image. 

Following existing works~\cite{cao2024monoscene, ruiqian2025splatssc}, our SSC loss function $\mathcal{L}_\mathrm{ssc}$ is a streamlined combination of focal loss, Lovasz loss, and geometry scale loss~\cite{li2017focalloss, maxim2018lovaszloss, cao2024monoscene}:
\begin{equation}
\mathcal{L}_\mathrm{ssc} = \lambda_1 \mathcal{L}_\mathrm{focal} + \lambda_2 \mathcal{L}_\mathrm{lovasz} + \mathcal{L}_\mathrm{scale}^\mathrm{geo}.
\end{equation}

To ensure feature-to-output consistency, the network is supervised on both GSE and DTE outputs. Since higher-level predictions are more important, we apply a decayed loss weight $w_j = \frac{2^j}{2^n -1}$ on $\mathcal{L}_\mathrm{ssc}$ for output layer $j$, where $n$ denotes the number of supervised outputs. The total objective is therefore expressed as
\begin{equation}
\mathcal{L}_{\mathrm{total}} 
= \sum_{j=1}^{n} w_{j} \mathcal{L}_{\mathrm{ssc}}^{j}.
\end{equation}
This stage aligns intermediate representations with voxel-level supervision 
and provides a well-initialized foundation for the subsequent embodied fine-tuning.

\noindent \textbf{Stage 2: Embodied Fine-tuning.}
The second stage adapts the pretrained monocular model to the embodied prediction task. Training samples are grouped by scene to preserve temporal continuity. During this phase, only the DTE is updated while the other components remain frozen. This selective optimization encourages the DTE to learn temporal fusion by aligning current and historical features in both geometry and semantics without disrupting the pretrained single-frame representations. This fine-tuning stage enables the model to perform temporal reasoning and make consistent occupancy predictions in embodied scenes.

 
\begin{table*}[t]
\centering
\setlength{\tabcolsep}{0.85mm}
\caption{\textbf{Local Prediction Performance on the Occ-ScanNet and Occ-ScanNet-mini dataset.} The best results are highlighted in \colorbox[HTML]{DEF3E6}{\textbf{bold}}, while the second-best are \colorbox[HTML]{ecf8f1}{\underline{underlined}}.}
\vspace{-8pt} 
\resizebox{0.98\textwidth}{!}{%
\begin{tabular}{l|c|c|c|ccccccccccc|c}
\toprule 
Dataset & Method & Input & IoU & \rotatebox{90}{\textcolor{ceiling}{\rule{0.6em}{0.6em}} ceiling} & \rotatebox{90}{\textcolor{floor}{\rule{0.6em}{0.6em}} floor} & \rotatebox{90}{\textcolor{wall}{\rule{0.6em}{0.6em}} wall} & \rotatebox{90}{\textcolor{window}{\rule{0.6em}{0.6em}} window} & \rotatebox{90}{\textcolor{chair}{\rule{0.6em}{0.6em}} chair} & \rotatebox{90}{\textcolor{bed}{\rule{0.6em}{0.6em}} bed} & \rotatebox{90}{\textcolor{sofa}{\rule{0.6em}{0.6em}} sofa} & \rotatebox{90}{\textcolor{table}{\rule{0.6em}{0.6em}} table} & \rotatebox{90}{\textcolor{tvs}{\rule{0.6em}{0.6em}} tvs} & \rotatebox{90}{\textcolor{furniture}{\rule{0.6em}{0.6em}} furniture} & \rotatebox{90}{\textcolor{objects}{\rule{0.6em}{0.6em}} objects} & mIoU \\

\midrule
\multirow{10}{*}{Occ-ScanNet} 
& TPVFormer~\cite{huang2023tri} & $x_t$ & 33.39 & 6.96 & 32.97 & 14.41 & 9.10 & 24.01 & 41.49 & 45.44 & 28.61 & 10.66 & 35.37 & 25.31 & 24.94 \\ 
& GaussianFormer~\cite{huang2024gaussianformer} & $x_t$ & 40.91 & 20.70 & 42.00 & 23.40 & 17.40 & 27.00 & 44.30 & 44.80 & 32.70 & 15.30 & 36.70 & 25.00 & 29.93 \\ 
& MonoScene~\cite{cao2024monoscene} & $x_t$ & 41.60 & 15.17 & 44.71 & 22.41 & 12.55 & 26.11 & 27.03 & 35.91 & 28.32 & 6.57 & 32.16 & 19.84 & 24.62 \\
& ISO~\cite{yu2024monocular} & $x_t$ & 42.16 & 19.88 & 41.88 & 22.37 & 16.98 & 29.09 & 42.43 & 42.00 & 29.60 & 10.62 & 36.36 & 24.61 & 28.71 \\
& SurroundOcc~\cite{wei2023surroundocc} & $x_t$ & 42.52 & 18.90 & 49.30 & 24.80 & 18.00 & 26.80 & 42.00 & 44.10 & 32.90 & 18.60 & 36.80 & 26.90 & 30.83 \\ 
& EmbodiedOcc~\cite{wu2024embodiedocc} & $x_t$ & 53.95 & 40.90 & 50.80 & 41.90 & 33.00 & 41.20 & 55.20 & 61.90 & 43.80 & 35.40 & 53.50 & 42.90 & 45.48 \\
& EmbodiedOcc++~\cite{wang2025embodiedoccplusplus} & $x_t$ & 54.90 & 36.40 & 53.10 & 41.80 & 34.40 & 42.90 & 57.30 & 64.10 & 45.20 & 34.80 & 54.20 & 44.10 & 46.20 \\
& RoboOcc~\cite{zhang2025roboocc} & $x_t$ & 56.48 & 45.36 & 53.49 & 44.35 & 34.81 & 43.38 & 56.93 & 63.35 & 46.35 & 36.12 & 55.48 & 44.78 & 47.67 \\
& SplatSSC~\cite{ruiqian2025splatssc} & $x_t$ & \cellcolor[HTML]{ecf8f1}\underline{62.83} & \cellcolor[HTML]{ecf8f1}\underline{49.10} & \cellcolor[HTML]{ecf8f1}\underline{59.00} & \cellcolor[HTML]{ecf8f1}\underline{48.30} & \cellcolor[HTML]{ecf8f1}\underline{38.80} & \cellcolor[HTML]{ecf8f1}\underline{47.40} & \cellcolor[HTML]{ecf8f1}\underline{62.40} & \cellcolor[HTML]{ecf8f1}\underline{67.00} & \cellcolor[HTML]{ecf8f1}\underline{49.50} & \cellcolor[HTML]{ecf8f1}\underline{42.60} & \cellcolor[HTML]{ecf8f1}\underline{60.70} & \cellcolor[HTML]{ecf8f1}\underline{45.40} & \cellcolor[HTML]{ecf8f1}\underline{51.83} \\ 
& TGSFormer (ours) & $x_t$ & \cellcolor[HTML]{DEF3E6}\textbf{64.42} & \cellcolor[HTML]{DEF3E6}\textbf{54.20} & \cellcolor[HTML]{DEF3E6}\textbf{61.30} & \cellcolor[HTML]{DEF3E6}\textbf{50.50} & \cellcolor[HTML]{DEF3E6}\textbf{41.30} & \cellcolor[HTML]{DEF3E6}\textbf{49.50} & \cellcolor[HTML]{DEF3E6}\textbf{64.50} & \cellcolor[HTML]{DEF3E6}\textbf{69.30} & \cellcolor[HTML]{DEF3E6}\textbf{52.40} & \cellcolor[HTML]{DEF3E6}\textbf{47.10} & \cellcolor[HTML]{DEF3E6}\textbf{63.00} & \cellcolor[HTML]{DEF3E6}\textbf{48.80} & \cellcolor[HTML]{DEF3E6}\textbf{54.73} \\
 
\midrule
\multirow{6}{*}{Occ-ScanNet-mini}
& MonoScene~\cite{cao2024monoscene} & $x_t$ & 41.90 & 17.00 & 46.20 & 23.90 & 12.70 & 27.00 & 29.10 & 34.80 & 29.10 & 9.70 & 34.50 & 20.40 & 25.90 \\ 
& ISO~\cite{yu2024monocular} & $x_t$ & 42.90 & 21.10 & 42.70 & 24.60 & 15.10 & 30.80 & 41.00 & 43.30 & 32.20 & 12.10 & 35.90 & 25.10 & 29.40 \\
& EmbodiedOcc~\cite{wu2024embodiedocc} & $x_t$ & 55.13 & 29.50 & 49.40 & 41.70 & 36.30 & 41.90 & 60.40 & 59.60 & 46.30 & 34.50 & 58.00 & 43.50 & 45.57 \\ 
& EmbodiedOcc++~\cite{wang2025embodiedoccplusplus} & $x_t$ & 55.70 & 23.30 & 51.00 & 42.80 & 39.30 & 43.50 & \cellcolor[HTML]{ecf8f1}\underline{65.60} & \cellcolor[HTML]{ecf8f1}\underline{64.00} & \cellcolor[HTML]{ecf8f1}\underline{50.70} & \cellcolor[HTML]{ecf8f1}\underline{40.70} & 60.30 & \cellcolor[HTML]{ecf8f1}\underline{48.90} & 48.20 \\
& SplatSSC~\cite{ruiqian2025splatssc} & $x_t$ & \cellcolor[HTML]{ecf8f1}\underline{61.47} & \cellcolor[HTML]{ecf8f1}\underline{36.60} & \cellcolor[HTML]{ecf8f1}\underline{55.70} & \cellcolor[HTML]{ecf8f1}\underline{46.50} & \cellcolor[HTML]{ecf8f1}\underline{40.10} & \cellcolor[HTML]{ecf8f1}\underline{45.60} & 64.50 & 62.40 & 48.60 & 30.60 & \cellcolor[HTML]{ecf8f1}\underline{61.20} & 45.39 & \cellcolor[HTML]{ecf8f1}\underline{48.87} \\
& TGSFormer (ours) & $x_t$ & \cellcolor[HTML]{DEF3E6}\textbf{66.19} & \cellcolor[HTML]{DEF3E6}\textbf{42.90} & \cellcolor[HTML]{DEF3E6}\textbf{62.30} & \cellcolor[HTML]{DEF3E6}\textbf{52.50} & \cellcolor[HTML]{DEF3E6}\textbf{47.10} & \cellcolor[HTML]{DEF3E6}\textbf{50.50} & \cellcolor[HTML]{DEF3E6}\textbf{70.30} & \cellcolor[HTML]{DEF3E6}\textbf{68.60} & \cellcolor[HTML]{DEF3E6}\textbf{56.90} & \cellcolor[HTML]{DEF3E6}\textbf{42.70} & \cellcolor[HTML]{DEF3E6}\textbf{67.80} & \cellcolor[HTML]{DEF3E6}\textbf{52.40} & \cellcolor[HTML]{DEF3E6}\textbf{55.82} \\

\bottomrule
\end{tabular}%
}
\vspace{-6pt}
\label{tab:occ_scannet_performance}
\end{table*}

\begin{table*}[!t] 	
\caption{\textbf{Embodied Prediction Performance on the EmbodiedOcc-ScanNet dataset.} The best results are highlighted in \colorbox[HTML]{DEF3E6}{\textbf{bold}}, while the second-best are \colorbox[HTML]{ecf8f1}{\underline{underlined}}.}
\vspace{-8pt}
\setlength{\tabcolsep}{0.85mm}
\centering
\resizebox{0.98\linewidth}{!}{
\begin{tabular}{l|c|c|c c c c c c c c c c c|c}
\toprule 
Method & Input & {IoU}
& \rotatebox{90}{\parbox{1.5cm}{\textcolor{ceiling}{$\blacksquare$} ceiling}} 
& \rotatebox{90}{\textcolor{floor}{$\blacksquare$} floor}
& \rotatebox{90}{\textcolor{wall}{$\blacksquare$} wall} 
& \rotatebox{90}{\textcolor{window}{$\blacksquare$} window} 
& \rotatebox{90}{\textcolor{chair}{$\blacksquare$} chair} 
& \rotatebox{90}{\textcolor{bed}{$\blacksquare$} bed} 
& \rotatebox{90}{\textcolor{sofa}{$\blacksquare$} sofa} 
& \rotatebox{90}{\textcolor{table}{$\blacksquare$} table} 
& \rotatebox{90}{\textcolor{tvs}{$\blacksquare$} tvs} 
& \rotatebox{90}{\textcolor{furniture}{$\blacksquare$} furniture} 
& \rotatebox{90}{\textcolor{objects}{$\blacksquare$} objects} 
& mIoU\\ 
\midrule 
TPVFormer~\cite{huang2023tri} & $\mathcal{X}$ & 35.88 & 1.62 & 30.54 & 12.03 & 13.22 & 35.47 & 51.39 & 49.79 & 25.63 & 3.60 & 43.15 & 16.23 & 25.70 \\
SurroundOcc~\cite{wei2023surroundocc} & $\mathcal{X}$ & 37.04 & 12.70 & 31.80 & 22.50 & 22.00 & 29.90 & 44.70 & 36.50 & 24.60 & 11.50 & 34.40 & 18.20 & 26.27 \\ 
GaussianFormer~\cite{huang2024gaussianformer} & $\mathcal{X}$ & 38.02 & 17.00 & 33.60 & 21.50 & 21.70 & 29.40 & 47.80 & 37.10 & 24.30 & 15.50 & 36.20 & 16.80 & 27.36 \\
EmbodiedOcc~\cite{wu2024embodiedocc} & $\mathcal{X}$ & 51.52 & 22.70 & \cellcolor[HTML]{DEF3E6}\textbf{44.60} & 37.40 & 38.00 & 50.10 & 56.70 & \cellcolor[HTML]{ecf8f1}\underline{59.70} & 35.40 & 38.40 & 52.00 & 32.90 & 42.53 \\
EmbodiedOcc++~\cite{wang2025embodiedoccplusplus} & $\mathcal{X}$ & 52.20 & 27.90 & 43.90 & 38.70 & \cellcolor[HTML]{DEF3E6}\textbf{40.60} & 49.00 & \cellcolor[HTML]{ecf8f1}\underline{57.90} & 59.20 & \cellcolor[HTML]{ecf8f1}\underline{36.80} & 37.80 & 53.50 & 34.10 & 43.60 \\
RoboOcc~\cite{zhang2025roboocc} & $\mathcal{X}$ & \cellcolor[HTML]{ecf8f1}\underline{53.30} & 21.94 & \cellcolor[HTML]{ecf8f1}\underline{44.57} & \cellcolor[HTML]{ecf8f1}\underline{39.54} & \cellcolor[HTML]{ecf8f1}\underline{38.48} & \cellcolor[HTML]{ecf8f1}\underline{51.28} & 57.04 & \cellcolor[HTML]{DEF3E6}\textbf{63.09} & 36.70 & \cellcolor[HTML]{DEF3E6}\textbf{43.05} & \cellcolor[HTML]{ecf8f1}\underline{54.42} & \cellcolor[HTML]{ecf8f1}\underline{34.38} & \cellcolor[HTML]{ecf8f1}\underline{44.05} \\
\midrule 

TGSFormer-C (ours) & $\mathcal{X}$ & 41.95 & \cellcolor[HTML]{ecf8f1}\underline{29.60} & 33.40 & 29.90 & 30.70 & 37.00 & 51.00 & 55.30 & 32.90 & 29.20 & 49.40 & 33.10 & 37.40 \\
TGSFormer (ours) & $\mathcal{X}$ & \cellcolor[HTML]{DEF3E6}\textbf{54.42} & \cellcolor[HTML]{DEF3E6}\textbf{31.00} & 39.90 & \cellcolor[HTML]{DEF3E6}\textbf{39.90} & 32.70 & \cellcolor[HTML]{DEF3E6}\textbf{52.00} & \cellcolor[HTML]{DEF3E6}\textbf{66.10} & 53.00 & \cellcolor[HTML]{DEF3E6}\textbf{47.30} & \cellcolor[HTML]{ecf8f1}\underline{38.70} & \cellcolor[HTML]{DEF3E6}\textbf{59.60} & \cellcolor[HTML]{DEF3E6}\textbf{37.90} & \cellcolor[HTML]{DEF3E6}\textbf{45.29} \\ 
\bottomrule
\end{tabular}
} 
\vspace{-12pt}
\label{tab:global_main}
\end{table*}

\section{Experiments}
We evaluate TGSFormer on both monocular and embodied semantic scene completion tasks. 
Monocular experiments are conducted on Occ-ScanNet and Occ-ScanNet-mini~\cite{yu2024monocular}, 
while embodied prediction is evaluated on EmbodiedOcc-ScanNet and EmbodiedOcc-ScanNet-mini~\cite{wu2024embodiedocc}.
Dataset details, implementation settings, and evaluation metrics are provided in the supplementary material.

\subsection{Main Result}
\noindent\textbf{Monocular Occupancy Prediction.} 
Tab.~\ref{tab:occ_scannet_performance} presents our experimental comparison with existing state-of-the-art (SOTA) methods on monocular SSC. 
On Occ-ScanNet, TGSFormer surpasses all compared methods by $1.59\%$ in geometry and $2.90\%$ in semantics. On the reduced Occ-ScanNet-mini, the margins further widen to $4.72\%$ and $6.95\%$. 
Notably, these gains hold across all semantic classes on both datasets, indicating that TGSFormer provides a robust monocular perception prior that scales well with varying dataset sizes.
We also provide further qualitative comparison with representative works in Fig.~\ref{fig:vis_base}. 

\noindent\textbf{Embodied Occupancy Prediction.} 
Tab.~\ref{tab:global_main} reports our main comparison on embodied SSC. 
On EmbodiedOcc-ScanNet, TGSFormer surpasses previous SOTA methods by $1.10\%$ in IoU and $1.20\%$ in mIoU, while using over $20\%$ fewer primitives (as reported in Fig.~\ref{fig:intro}).
For class-wise performance, TGSFormer outperforms all compared methods on 7/11 semantic categories, with slight drops in large uniform regions due to feature smoothing. 
Similar to prior works~\cite{wu2024embodiedocc,wang2025embodiedoccplusplus, zhang2025roboocc}, we include an additional baseline, TGSFormer-C, which aggregates local predictions and directly concatenates loaded historical Gaussians. 
Despite its strong performance in the monocular setting, TGSFormer-C degrades notably in the embodied scenario, highlighting the necessity of a principled memory maintenance mechanism. 
As illustrated in Fig.~\ref{fig:vis_glob}, our TGSFormer progressively expands the explored area and refines previously observations, producing globally consistent reconstructions, even under frame-agnostic scenarios. 

\subsection{Ablation Studies}
In this section, we conduct ablation studies on the key modules proposed in our framework 
to comprehensively verify the effectiveness of each component.

\begin{figure}[!t]
\centering 
\vspace{-8pt}
\includegraphics[width=0.99\columnwidth]{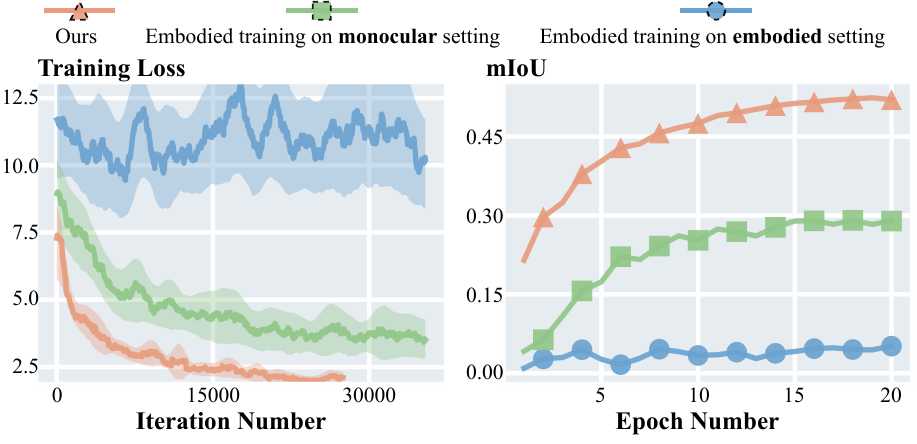}
\vspace{-10pt}
\caption{\textbf{Comparison of Training Strategy}.} 
\vspace{-4pt}
\label{fig:training_strategy} 
\end{figure}

\noindent\textbf{Training Strategy Comparison.} We compare our two-stage training strategy with directly training the temporal model. As shown in Fig.~\ref{fig:training_strategy}, monocular first-stage training converges faster, runs more stably, and yields higher performance in mIoU than training the temporal model alone. This behavior is consistent with our training objective in Sec.~\ref{sec:training_objective}, where temporally independent frames are used to establish a scene-agnostic perceptual prior.

\noindent\textbf{Ablation on CAVF module.} We evaluate the effectiveness of the proposed Confidence-aware Voxel Fusion module. 
We set different voxel sizes to analyze how voxel granularity influences the final reconstruction performance, 
and further examine the impact of incorporating confidence information during fusion. 
The results are shown in Tab.~\ref{tab:ablation_gaussian_merging}. Our design achieves consistent accuracy improvements while significantly reducing the memory requirements 
of the scene representation.

\begin{table}[t]
\centering 
\setlength{\tabcolsep}{1mm} 
\caption{\textbf{Ablation on CAVF}. The results of our proposed setting are highlighted in light \colorbox{gray!15}{gray}.}
\vspace{-8pt} 
\resizebox{0.95\columnwidth}{!}{%
\begin{tabular}{c|cc|cc|cc}
\toprule 
CAVF & Voxel Size & Conf. & \makecell{Memory\\(MiB)} & \makecell{Number\\(Per-frame)} & IoU$\uparrow$ & mIoU$\uparrow$ \\
\midrule
 / & / & / & 3696 & 4086.0 & 60.48 & 41.84 \\ 
\checkmark  & 0.12m & / & 3322 & 876.1 & 59.76 & 45.76 \\
\checkmark  & 0.08m & \checkmark & 3328 & 1507.6 & 64.18 & 49.61 \\
\checkmark  & 0.10m & \checkmark & 3324 & 1124.3 & \textbf{64.35} & \underline{50.38} \\ 
\cellcolor{gray!15}\checkmark  & \cellcolor{gray!15}0.12m & \cellcolor{gray!15}\checkmark & \cellcolor{gray!15}3320 & \cellcolor{gray!15}859.0 & \cellcolor{gray!15}\underline{64.16} & \cellcolor{gray!15}\textbf{50.55} \\ 
\checkmark  & 0.14m & \checkmark & 3322 & 671.0 & 63.45 & 49.88 \\ 
\bottomrule
\end{tabular}%
}
\vspace{-6pt}
\label{tab:ablation_gaussian_merging}
\end{table} 

\noindent \textbf{Training Objective and Feature  Alignment.} 
We conduct an ablation study on different supervision strategies and evaluate the results on both the monocular and embodied tasks, as shown in Tab.~\ref{tab:multi_stage_supervise}.
When supervision is applied only at Stage~2, the results are suboptimal because the intermediate features remain poorly aligned with the history Gaussians. However, applying supervision across all stages also leads to degraded performance, as overly dense constraints over-regularize the feature hierarchy. The best results on embodied settings are achieved when supervision is applied at Stages~1 and~2, suggesting that moderate mid-level supervision provides sufficient guidance for progressive feature alignment without disrupting representation learning.

\begin{table}[t]
\centering
\setlength{\tabcolsep}{1mm}
\caption{\textbf{Ablation on Training Objective and Feature Alignment.}We validate our loss components and our multi-stage supervision strategy for feature alignment. The \textbf{Stages} column refers to supervising the outputs of the GSE (0,1) and DTE (2).}
\vspace{-6pt}
\resizebox{0.95\columnwidth}{!}{%
\begin{tabular}{cc|cc|cc}
\toprule
\multicolumn{2}{c|}{\textbf{Supervision Strategy}} & \multicolumn{2}{c|}{\textbf{Monocular Setting}} & \multicolumn{2}{c}{\textbf{Embodied Setting}} \\ 
Training Objective & Stages & IoU $\uparrow$ & mIoU $\uparrow$ & IoU $\uparrow$ & mIoU $\uparrow$ \\
\midrule
$\mathcal{L}_{\mathrm{focal}}, \mathcal{L}_{\mathrm{lovasz}}, \mathcal{L}_{\mathrm{prob}}$ & / & \underline{64.54} & \textbf{54.62} & 61.75 & 47.21 \\
$\mathcal{L}_{\mathrm{focal}}, \mathcal{L}_{\mathrm{lovasz}}, \mathcal{L}_{\mathrm{geo}}$  & 0, 1, 2 & 64.11 & 53.69 & \underline{62.20} & \underline{48.63} \\
\cellcolor{gray!15}$\mathcal{L}_{\mathrm{focal}}, \mathcal{L}_{\mathrm{lovasz}}, \mathcal{L}_{\mathrm{geo}}$ & \cellcolor{gray!15}1, 2 & \cellcolor{gray!15}\textbf{64.73} & \cellcolor{gray!15}54.41 & \cellcolor{gray!15}\textbf{63.31} & \cellcolor{gray!15}\textbf{49.70} \\
$\mathcal{L}_{\mathrm{focal}}, \mathcal{L}_{\mathrm{lovasz}}, \mathcal{L}_{\mathrm{geo}}$  & 2 & 64.39 & \underline{54.54} & 61.26 & 48.04 \\ 
\bottomrule 
\end{tabular}%
}
\vspace{-10pt}
\label{tab:multi_stage_supervise}
\end{table}

\noindent \textbf{Gaussian initialization and temporal encoder.}
We perform ablation studies on both the Gaussian initialization and the cross-attention mechanisms to validate the effectiveness of our proposed design. 
The results indicate that our gains do not merely stem from the strong depth features provided by DepthAnythingV2. For fairness, depth-based baselines~\cite{wu2024embodiedocc, wang2025embodiedoccplusplus, zhang2025roboocc} use the same estimator, and across both strong and weaker depth models, TGSFormer consistently achieves the best performance.
Furthermore, the ablation on the cross-attention configuration confirms the effectiveness of our proposed dual-attention module and confidence modulation, which together contribute to more accurate and stable fusion of geometric and appearance cues. 

\begin{table}[t]
\centering
\footnotesize
\setlength{\tabcolsep}{1mm}
\caption{\textbf{Ablation on Gaussian Initialization and Temporal Encoder.} (a) Different settings for depth cue. (b) Study on different temporal processing strategies. Including the baseline without temporal modeling, single cross attention (ca), and dual cross attention (dual ca) variants, with or without confidence modulation.}
\vspace{-8pt}
\begin{subtable}[t]{0.40\linewidth}
    \centering
    \begin{tabular}{l|cc}
        \toprule
        Initialize & IoU $\uparrow$ & mIoU $\uparrow$ \\
        \midrule
        G. T. & \textbf{68.21} & \textbf{56.94} \\
        DaV2 & 50.15 & 39.50 \\
        \cellcolor{gray!15}FT. DaV2 & \cellcolor{gray!15}\underline{64.74} & \cellcolor{gray!15}\underline{54.41} \\
        DA~\cite{wu2024embodiedocc} & 64.49 & 54.34 \\
        GMF~\cite{ruiqian2025splatssc} & 60.71 & 49.02 \\
        \bottomrule
    \end{tabular}
    \caption{Gaussian Initialization.}
    \label{tab:ablation_gaussian_initialization}
\end{subtable}
\hfill
\begin{subtable}[t]{0.58\linewidth}
    \centering
    \begin{tabular}{lc|ll}
        \toprule 
        \multicolumn{1}{c}{Temporal} & \multicolumn{1}{c|}{Conf.} & \multicolumn{1}{c}{IoU $\uparrow$} & \multicolumn{1}{c}{mIoU $\uparrow$} \\
        \midrule 
        / & / & 61.64 & \underline{49.35} \\ 
        ca & / & 60.37 \scriptsize-\tiny1.27 & 48.52 \scriptsize-\tiny0.83 \\
        ca & \checkmark & 60.92 \scriptsize-\tiny0.72 & 48.72 \scriptsize-\tiny0.83 \\
        dual ca & / & \underline{62.77} \scriptsize+\tiny1.13 & 48.29 \scriptsize-\tiny1.06 \\
        \cellcolor{gray!15}dual ca & \cellcolor{gray!15}\checkmark & \cellcolor{gray!15}\textbf{63.31} \textbf{\scriptsize+\tiny1.67} & \cellcolor{gray!15}\textbf{49.70} \textbf{\scriptsize+\tiny0.35} \\
        \bottomrule
    \end{tabular} 
    \caption{Temporal Encoder Strategies.}
    \label{tab:ablation_dual_temporal_encoder}
\end{subtable}
\vspace{-10pt}
\end{table}

\noindent \textbf{CCA modulation strategies.}
For the ablation study on the CCA module, we compare results of integrating the confidence modulation at different positions within the network. 
As shown in Tab.~\ref{tab:ablation_modulation}, the best performance is achieved when the confidence module is applied at the positions $C_v$ and $C_a$, 
as illustrated in Fig.~\ref{fig:modulation-vis}.
This configuration yields the best joint improvement in both geometry and semantics.

\begin{table}[t] 
\centering
\footnotesize
\caption{ 
    \textbf{Ablation on CCA modulation strategies}. 
    We compare the confidence modulation on the query ($C_q$), value ($C_v$), and concatenated attention output ($C_a$).
} 
\vspace{-8pt}
\adjustbox{valign=c}{ 
\begin{subtable}{0.40\linewidth} %
    \centering
    \setlength{\tabcolsep}{0.01\linewidth}
    \begin{tabular}{l|ll} 
        \toprule
        \multicolumn{1}{c|}{Modulate} & \multicolumn{1}{c}{IoU $\uparrow$} & \multicolumn{1}{c}{mIoU $\uparrow$} \\
        \midrule
        / & 64.28 & \underline{54.03} \\ 
        $C_q$ & 64.15 \scriptsize-\tiny0.13 & 53.55 \scriptsize-\tiny0.48 \\
        $C_{a}$ & 64.26 \scriptsize-\tiny0.02 & 53.89 \scriptsize-\tiny0.14 \\
        $C_{v}$ & \underline{64.43} \scriptsize+\tiny0.15 & 53.98 \scriptsize-\tiny0.05 \\
        \cellcolor{gray!15}$C_v, C_a$ & \cellcolor{gray!15}\textbf{64.74} \textbf{\scriptsize+\tiny0.46} & \cellcolor{gray!15}\textbf{54.41} \textbf{\scriptsize+\tiny0.38} \\
        \bottomrule
    \end{tabular}
    \caption{Modulation strategies.}
    \label{tab:ablation_modulation}
\end{subtable}
}
\hfill
\adjustbox{valign=c}{
\begin{subtable}{0.55\linewidth} %
    \centering
    \includegraphics[width=\linewidth]{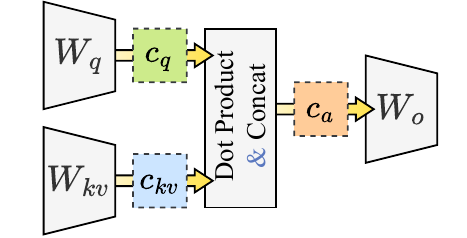}
    \caption{Modulate position illustration.}
    \label{fig:modulation-vis}
\end{subtable}
}
\vspace{-16pt}
\end{table}

\section{Conclusion}
We presented TGSFormer, a scalable temporal Gaussian splatting framework for embodied semantic scene completion. 
By maintaining a persistent Gaussian memory with confidence-aware temporal and voxel fusion, TGSFormer achieves compact and high-fidelity reconstruction across long-term embodied exploration. 
Extensive experiments demonstrate clear advantages in scalability, accuracy, and efficiency on both monocular and embodied benchmarks, 
establishing a solid foundation for future research in large-scale memory-driven 3D perception. 
Detailed discussions and limitations are provided in the supplementary material.

\clearpage
\setcounter{page}{1}
\maketitlesupplementary

\section{Experimental Setup}
\subsection{Dataset} 
\noindent \textbf{Occ-ScanNet.} 
Occ-ScanNet~\cite{yu2024monocular} is a large-scale indoor monocular semantic occupancy dataset containing 45,755 training frames and 19,764 validation frames. Each sample is annotated with 12 semantic categories, including free space and eleven occupied classes: ceiling, floor, wall, window, chair, bed, sofa, table, television, furniture, and generic objects.
The ground-truth occupancy is provided as a voxel grid covering a 4.8m × 4.8m × 2.88m region in front of the camera, discretized into a 60 × 60 × 36 resolution.
This dataset is used as the benchmark for our local monocula occupancy prediction experiments. We also report results on Occ-ScanNet-mini, a reduced subset containing 5,504 training frames and 2,376 validation frames.
 
\noindent \textbf{EmbodiedOcc-ScanNet.}  
EmbodiedOcc-ScanNet~\cite{wu2024embodiedocc} provides an embodied variant of ScanNet, consisting of 537 training scenes and 137 validation scenes. Each scene contains a short exploration sequence of 30 posed RGB images together with corresponding volumetric occupancy ground truth.
The global occupancy for each scene is generated by voxelizing the entire traversed region in world coordinates, using the same voxel size and semantic label set as the local task. In the embodied setting, the model receives sequential observations and updates a global scene estimate conditioned on the known camera poses. 

\subsection{Temporal-Occ-ScanNet Variant.}
As Occ-ScanNet lacks sequential exploration, we additionally construct a temporalized variant of the dataset, termed Temporal-Occ-ScanNet for clarity. Beyond enabling controlled evaluation of our two-stage training strategy, this set also offers a lightweight benchmark that still contains sufficient temporal complexity for analyzing temporal fusion behaviours. Concretely, we reorganize all frames by scene, divide each scene into fixed-length \textbf{batches} that preserve intra-scene continuity, and enqueue the batches in sequential order. Frames within the same batch therefore form short yet meaningful temporal windows.

To ensure that each batch contains sufficient temporal context, we apply a scene-consistent padding strategy: if a batch contains fewer frames than the target temporal length, we automatically pad it using neighboring frames from the same scene, prioritizing earlier frames, followed by later ones, and finally repeating existing frames when necessary. This guarantees smooth temporal transitions even for scenes with limited frame counts.

During training, batches are shuffled at the group level rather than at the frame level, preserving temporal coherence within each batch while still exposing the model to diverse scene orders. Each data sample thus provides a compact sequence of RGB images, depth maps, intrinsic/extrinsic parameters, and voxel occupancies aligned in time, enabling fair and stable temporal supervision on Occ-ScanNet without altering its original annotations.

\subsection{Evaluation metrics}
We evaluate semantic scene completion performance using Intersection-over-Union (IoU) and mean IoU (mIoU) over the 12 semantic categories.

For local occupancy prediction, we follow the evaluation protocol of ISO~\cite{yu2024monocular}, computing IoU strictly within the current frame’s camera frustum.

For embodied occupancy prediction, we adhere to the EmbodiedOcc~\cite{wu2024embodiedocc} protocol. The evaluation is performed over the global voxel grid of each scene, considering only regions that are observed at least once throughout the 30-frame exploration sequence.

\begin{table*}[t] 
\small
\centering
\setlength{\tabcolsep}{1mm}
\caption{\textbf{Experiment settings for different ablation studies and efficient analysis.} Experiments above the dashed line are included in our main manuscript, while those below the dashed line are newly introduced in this appendix.}
\vspace{-6pt}
\resizebox{0.99\linewidth}{!}{
\begin{tabular}{l|cccc} 
\toprule
\multirow{2}{*}{Experiments} & \multicolumn{4}{c}{\textbf{Experiment Settings}} \\ 
& Training Dataset & Training Device & Max Learning Rate & Total Batch Size \\
\midrule 
Comparison of Training Strategy & Occ-ScanNet-mini & 2 NVIDIA RTX $3090$ & $6\times 10^{-4}$ & 6 \\
Ablation on CAVF & Temporal-Occ-ScanNet-mini & 4 NVIDIA RTX $3090$ & $4\times 10^{-4}$ & 4 \\
Ablation on Training Objective & Temporal-Occ-ScanNet-mini & 4 NVIDIA RTX $3090$ & $4\times 10^{-4}$ & 4 \\ 
Ablation on Gaussian Initialization & Occ-ScanNet-mini & 2 NVIDIA RTX $3090$ & $6\times 10^{-4}$ & 6 \\
Ablation on Temporal Encoder & Temporal-Occ-ScanNet-mini & 4 NVIDIA RTX $3090$ & $4\times 10^{-4}$ & 4 \\
Ablation on CCA modulation strategies & Occ-ScanNet-mini & 2 NVIDIA RTX $3090$ & $6\times 10^{-4}$ & 6 \\
\noalign{\vskip -5pt}
\multicolumn{5}{c}{\hdashrule{0.99\linewidth}{0.5pt}{2pt}}\\
\noalign{\vskip -1pt}
Ablation on Uncertainty Estimation & Occ-ScanNet-mini & 2 NVIDIA RTX $3090$ & $6\times 10^{-4}$ & 6 \\
Ablation on Confidence-aware Loss & Occ-ScanNet-mini & 2 NVIDIA RTX $3090$ & $6\times 10^{-4}$ & 6 \\
Efficiency Analysis & EmbodiedOcc-ScanNet-mini & 4 NVIDIA RTX $3090$ & $4\times 10^{-4}$ & 4 \\
\bottomrule 
\end{tabular} 
}
\vspace{-10pt}
\label{tab:ablation_config} 
\end{table*}
 
\subsection{Implementation Details}
In our framework, the image encoder employs a pretrained EfficientNet-B7~\cite{tan2019efficientnet} as backbone, while the depth branch utilizes a frozen fine-tuned \textit{Depth-Anything-V2}~\cite{wu2024embodiedocc} model. 

\noindent \textbf{Stage~1: Monocular Pretraining.}
In the first stage, we train TGSFormer on monocular SSC to establish a strong and frame-agnostic perceptual prior. The Gaussian Lifter operates on a uniformly downsampled grid of $30\times 40$ points, following SplatSSC~\cite{ruiqian2025splatssc}. 
For Confidence-aware Voxel Fusion (CAVF), we set the sharpness parameter $p$ and the maximum entropy threshold $H_\mathrm{max}$ to $3.0$. 
The loss weights $\lambda_1$ and $\lambda_2$ in the final objective $\mathcal{L}_\mathrm{total}$ are set to $100$ and $2$, respectively.  
We use the AdamW optimizer~\cite{loshchilov2017decoupled} with a weight decay of 0.01, and apply a learning-rate multiplier of 0.1 to the image backbone. The learning rate follows a cosine schedule with a 1000-iteration warmup, reaching a peak value of $8 \times 10^{-4}$. The model is trained for 10 epochs on Occ-ScanNet and for 20 epochs on Occ-ScanNet-mini, using 4 NVIDIA RTX 3090 GPUs with a batch size of 2 per GPU (global batch size 8).

\noindent \textbf{Stage~2: Embodied Fine-tuning.}
In the second stage, we adapt the pretrained model to the embodied setting. 
To preserve the frame-agnostic perceptual prior established in Stage~1, all components of TGSFormer are frozen, except for the Dual Temporal Encoder (DTE), which is exclusively responsible for temporal fusion. 
During finetuning, we apply a learning-rate multiplier of 0.1 to the DTE parameters, while the remainder of the network remains fixed. This selective optimization enables the model to learn stable cross-frame interactions and temporal consistency without perturbing the underlying single-frame representation.  
Stage~2 is trained for 5 epochs on EmbodiedOcc-ScanNet using 4 NVIDIA RTX 3090 GPUs, with a batch size of 1 per GPU (global batch size 4).

\noindent \textbf{Further experimental settings.}
The configurations used in our ablation studies and efficiency analyses are summarized in Tab.~\ref{tab:ablation_config}. Each experiment follows the same training and inference protocol as its corresponding main result, with changes applied only to the component being examined. All experiments are conducted on a single NVIDIA RTX 3090 GPU, and the inference dataset is identical to the training dataset unless otherwise specified.


\section{Further Experiment Results}

\begin{table}[t]
\centering 
\setlength{\tabcolsep}{1mm} 
\caption{\textbf{Ablation on Uncertainty Estimation.} 
The temperature parameter used in the normalization step is denoted as $T$. 
Among all variants, the power transform with $T=0.2$ achieves the best performance. 
The results of our proposed setting are highlighted in light \colorbox{gray!15}{gray}.}
\vspace{-8pt}
\resizebox{0.95\columnwidth}{!}{%
\begin{tabular}{ccc|cc}
\toprule
Uncertainty Transform & Normalize & Temperature & IoU$\uparrow$ & mIoU$\uparrow$ \\
\midrule
sharp sigmoid & softmax & 1.0 & 64.19 & 51.94 \\ 
power transform  & softmax & 1.0 & 64.20 & \underline{52.04} \\
\cellcolor{gray!15}power transform  & \cellcolor{gray!15}softmax & \cellcolor{gray!15}0.2 & \cellcolor{gray!15}\textbf{64.32} & \cellcolor{gray!15}\textbf{52.10} \\
power transform  & softmax & 0.5 & \underline{64.23} & 51.99 \\
\bottomrule
\end{tabular}%
} 
\vspace{-10pt}
\label{tab:ablation_certainty_estimation}
\end{table}

\subsection{Uncertainty Estimation} 
We further study the effect of different uncertainty-to-confidence mappings used in CAVF, as summarized in Tab.~\ref{tab:ablation_certainty_estimation}. Among all variants, the power transform with a temperature of $0.2$ achieves the highest IoU and mIoU, indicating that a stronger contrast in confidence weighting leads to more reliable voxel-level fusion. In comparison, the sharp sigmoid baseline is:
\begin{align}
    c = \sigma\big(-\beta\,(H-\gamma)\big),
\end{align}
where $H$ denotes the entropy and $(\beta,\gamma)$ are set to $(10.0, 1.5)$ in our experiments. Utilizing the sharp sigmoid leads to inferior performance on both IoU and mIoU compared to our power transform, since its steep nonlinearity pushes confidence toward near-binary values, making the fusion less stable. In contrast, our power transform offers smoother scaling, with a lower temperature gives marginally more reliable weights, yielding the best result.

\begin{figure}[!t]  
\centering
\includegraphics[width=0.95\columnwidth]{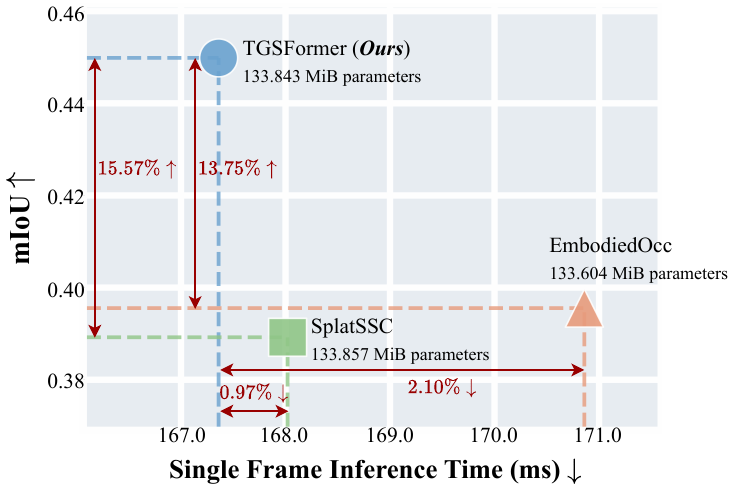}
\vspace{-8pt} 
\caption{
\textbf{Single-frame inference latency versus mIoU in Embodied Sequence}. 
All models share comparable parameter counts. 
EmbodiedOcc incurs the highest latency due to accumulating Gaussian features and memory entries. 
TGSFormer achieves both the lowest latency and the highest mIoU, benefiting from its lightweight Gaussian Lifter, DTE, and CAVF modules.
}
\label{fig:latency_estimate} 
\vspace{-10pt}
\end{figure} 

\begin{figure*}[!t]
\centering 
\vspace{-10pt}
\includegraphics[width=0.98\textwidth]{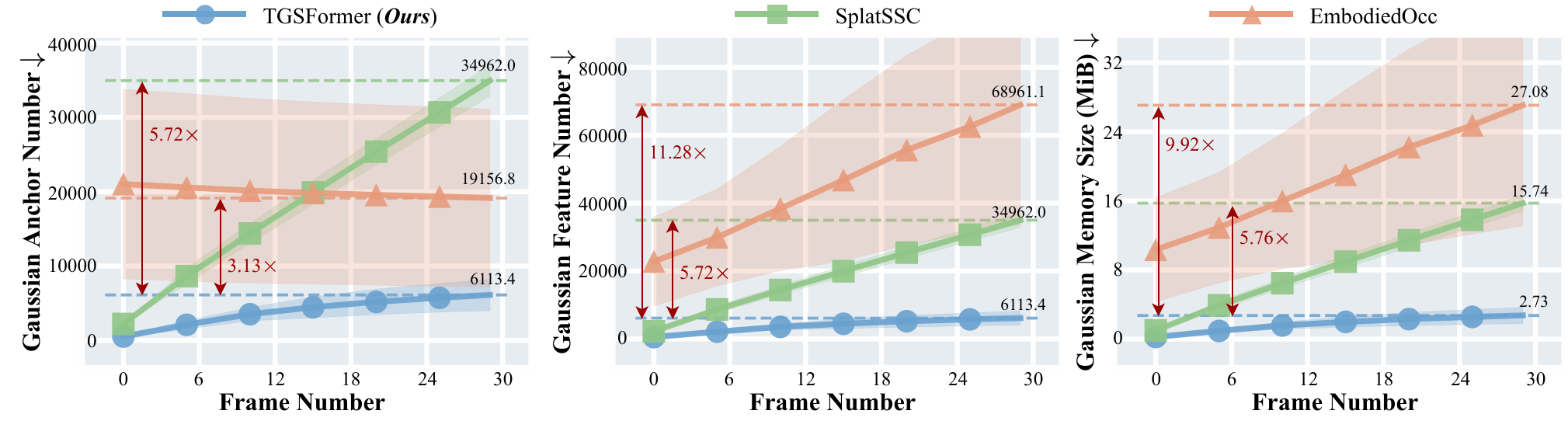} %
\vspace{-10pt}
\caption{ 
\textbf{Efficiency comparison in Embodied Sequence.} 
EmbodiedOcc shows unbounded growth in Gaussian features and memory, while SplatSSC exhibits steady accumulation due to the lack of temporal regulation. 
TGSFormer maintains a compact and bounded Gaussian representation through CAVF, resulting in up to $11.28\times$ and $9.92\times$ reductions in feature count and memory size, respectively.
}
\label{fig:efficiency_estimate}
\vspace{-8pt}
\end{figure*} 

\subsection{Efficiency Analysis}
\noindent \textbf{Runtime Efficiency.}
We evaluate the single-frame inference latency of TGSFormer, SplatSSC, and EmbodiedOcc in Fig.~\ref{fig:latency_estimate}. 
All three models share comparable parameter counts, ensuring a fair comparison. 
EmbodiedOcc exhibits the highest latency due to the continual growth of its Gaussian features and memory entries, which increases splatting and aggregation cost. 
SplatSSC achieves faster inference but remains slightly slower than TGSFormer, consistent with its more complex multi-branch feature fusion design and the gradual accumulation of Gaussian features over long sequences.

Despite incorporating both temporal fusion and memory regulation, 
TGSFormer attains the \textit{lowest} single-frame latency while simultaneously achieving the \textit{highest} mIoU. 
This efficiency stems from two design choices: (1) a simple depth-guided Gaussian Lifter without heavy geometric modules, 
and (2) lightweight Dual Temporal Encoder (DTE) and CAVF modules that introduce negligible computational overhead. 
Moreover, CAVF actively reduces the number of active Gaussians during the update process, further lowering per-frame splatting and rendering cost.

\noindent \textbf{Gaussian Complexity and Memory Growth.}
We further analyze the evolution of Gaussian anchors, features, and memory consumption throughout the embodied sequence in Fig.~\ref{fig:efficiency_estimate}. 
EmbodiedOcc initializes a moderate number of Gaussian anchors but lacks a mechanism to constrain the associated feature representations. 
Consequently, while the anchor count remains relatively stable across frames, its Gaussian features and memory usage grow rapidly due to unbounded accumulation.
SplatSSC exhibits similar growth in anchors and features, as it performs frame-wise depth-guided lifting without temporal regulation.

In contrast, TGSFormer maintains substantially fewer Gaussian anchors, features, and memory entries throughout the sequence. 
The CAVF module merges overlapping Gaussians in a confidence-aware manner, producing a compact and bounded representation that converges within only a few frames. 
Compared to EmbodiedOCC and SplatSSC, TGSFormer achieves up to \textbf{5.72$\times$ fewer anchors},  \textbf{11.28$\times$ fewer features}, and \textbf{9.92$\times$ less memory}.

\begin{figure}[!t]  
\centering
\includegraphics[width=0.98\columnwidth]{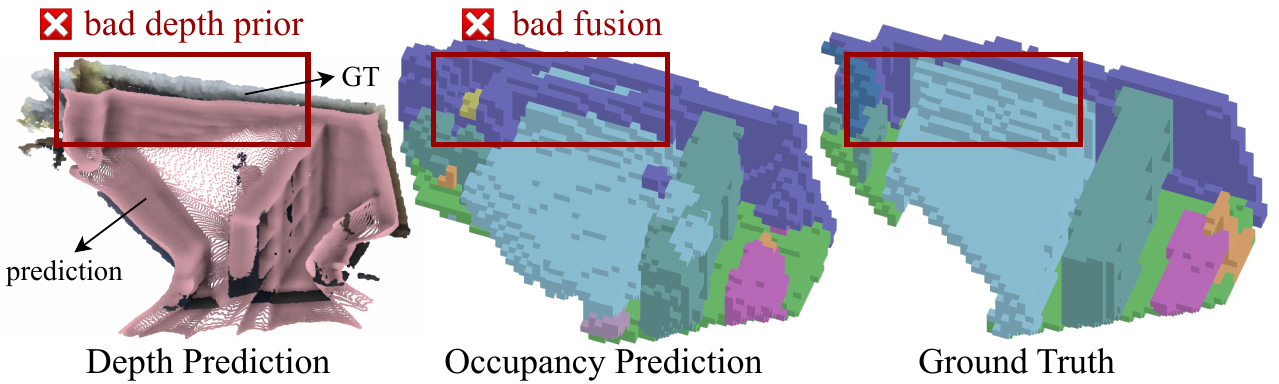} 
\vspace{-6pt} 
\caption{
\textbf{Failure case on Temporal-Occ-ScanNet-mini.} Large depth errors under severe occlusion (GT in color vs.\ prediction in pink) lead to misaligned lifted Gaussians, causing inconsistent occupancy between the current frame and the historical estimate.
}
\label{fig:failure} 
\vspace{-10pt} 
\end{figure} 

\section{Discussions} 
\subsection{Failure Example}
When the depth prior contains large metric-scale errors under severe occlusion, the lifted Gaussians of the current frame are misplaced far from the true geometry. Such errors cannot be corrected by the Dual Temporal Encoder (DTE), which operates under the assumption of locally consistent geometry, nor by the Confidence-aware Voxel Fusion (CAVF), whose merging radius is restricted to nearby voxel neighborhoods. As a result, the misaligned Gaussians persist through temporal fusion and lead to incorrect global occupancy in the embodied setting. This phenomenon is shown in the Fig.~\ref{fig:failure}, where inaccurate depth prediction (pink points) deviates significantly from the ground-truth structure (rgb points), causing visible inconsistency between the current and historical occupancy estimates.

\begin{table}[t]
\centering
\setlength{\tabcolsep}{1mm}
\caption{\textbf{Preliminary study on Confidence-aware Loss Strategies.}
Each method adds an extra confidence dimension to Gaussians and renders it by confidence-weighted splatting. 
During training, the confidence $C'_i$ is used to reweight the target loss with a weighting term parameterized by $\lambda$.}
\vspace{-6pt} 
\resizebox{0.97\columnwidth}{!}{
\begin{tabular}{c c c | c c}
\toprule
Reweighted Terms & Reweighting Formula & $C'_i$ Mapping & IoU$\uparrow$ & mIoU$\uparrow$ \\
\midrule 
/ & / & / & \textbf{63.13} & \textbf{52.89} \\ 
$\mathcal{L}_{\text{focal}}$ & $C'_i\mathcal{L}_{\text{target}}+\lambda e^{-C'_i}$ & $\exp(c'_i)$ & \underline{62.19} & \underline{52.12} \\
$\mathcal{L}_{\text{geo}}$ & $C'_i\mathcal{L}_{\text{target}}+\lambda e^{-C'_i}$ & $\exp(c'_i)$ & 61.81 & 51.86 \\
$\mathcal{L}_{\text{focal}},\mathcal{L}_{\text{geo}}$ & $C'_i\mathcal{L}_{\text{target}}+\lambda e^{-C'_i}$ & $\exp(c'_i)$ & 60.47 & 51.91 \\
$\mathcal{L}_{\text{focal}},\mathcal{L}_{\text{geo}}$ & $C'_i\mathcal{L}_{\text{target}}-\lambda\log(C'_i)$ & $1+\exp(c'_i)$ & 58.95 & 51.22 \\
\bottomrule
\end{tabular}
}
\vspace{-10pt}
\label{tab:confidence_loss}
\end{table}

\subsection{Future Works} 
\noindent \textbf{Uncertainty Quantification.} 
Our current confidence estimation is primarily based on semantic entropy and does not explicitly account for geometric uncertainty arising from depth errors or lifting drift. 
A more principled treatment of model and data uncertainty may benefit both temporal alignment and voxel fusion. 
Classical approaches such as MC dropout~\cite{gal2016dropout}, deep ensembles~\cite{lakshminarayanan2017simple}, and MIMO~\cite{havasi2021training} offer generic mechanisms for estimating epistemic and aleatoric uncertainty, and represent natural directions for extending our framework. 
Recent SSC works~\cite{cao2024pasco, severin2025occuq} incorporate uncertainty into 3D occupancy reasoning but are typically built upon SurroundOcc-style pipelines~\cite{wei2023surroundocc}, which do not directly align with our Gaussian-memory representation. 
EmbodiedOcc++~\cite{wang2025embodiedoccplusplus} adopts MC dropout for semantic uncertainty, yet still faces challenges in balancing efficiency and reconstruction performance. 

Another paradigm~\cite{wang2024dust3r, wang2025vggt} utilize confidence as an auxiliary prediction to weight the loss, rather than as a directly supervised input to fusion. Inspired by this, we added an additional confidence attribute $c'_i \in [0, 1]$ to each Gaussian primitive and rendered it by confidence-weighted splatting:
\begin{align}
\hat{C}(\mathbf{x}) = \frac{\sum_{i\in\mathcal{N}(\mathbf{x})} p(\mathbf{x}\mid G_i)\,c'_i}{\sum_{j\in\mathcal{N}(\mathbf{x})} p(\mathbf{x}\mid G_j)}. 
\end{align} 
Nevertheless, directly using this confidence to reweight the training loss (Tab.~\ref{tab:confidence_loss}) led to a slight drop in performance.
In particular, the formulation $C'_i\mathcal{L}_{\text{target}}-\lambda\log(C'_i)$ with $C'_i=1+\exp(c'_i)$ follows the confidence-aware loss design of Dust3R~\cite{wang2024dust3r}, while the alternative $C'_i\mathcal{L}_{\text{target}}+\lambda e^{-C'_i}$ with $C'_i=\exp(c'_i)$ is adopted to avoid negative loss values. 
Both variants underperform in our setting, indicating that loss reweighting schemes to Gaussian–based occupancy prediction are non-trivial and require further adaptation.

Designing unified semantic and geometric uncertainty estimators that integrate effectively with Gaussian memory remains an important direction for future work.
 
\noindent \textbf{RNN-style Scene Reconstruction.}
As shown in Fig.~\ref{fig:efficiency_estimate}, the number of Gaussian primitives in TGSFormer still grows over time in embodied settings, albeit much more slowly and non-linearly compared to existing baselines. 
This suggests that maintaining long-term memory consistency remains a challenge under extended exploration. 
Recent works~\cite{wang2025cut3r, chen2025ttt3r} demonstrate the feasibility of recurrent or state-space formulations for sustaining global 3D consistency over long sequences. 
Exploring RNN-style mechanisms for managing Gaussian memory and mitigating long-horizon drift represents a natural extension of this work.

{
    \small
    \bibliographystyle{configs/ieeenat_fullname}
    \bibliography{main}

@String(CVPR= {IEEE Conf. Comput. Vis. Pattern Recog.})

@String(ICCV= {Int. Conf. Comput. Vis.})

@String(ECCV= {Eur. Conf. Comput. Vis.})

@String(ICLR = {Int. Conf. Learn. Represent.})

@String(CVPR  = {CVPR})

@String(ICCV  = {ICCV})

@String(ECCV  = {ECCV})

@String(ICLR  = {ICLR})

@inproceedings{huang2023tri,
  author       = {Yuanhui Huang and
                  Wenzhao Zheng and
                  Yunpeng Zhang and
                  Jie Zhou and
                  Jiwen Lu
                 },
  title        = {Tri-Perspective View for Vision-Based 3D Semantic Occupancy Prediction},
  booktitle    = {Proceedings of the {IEEE/CVF} Conference on Computer Vision and Pattern Recognition},
  pages        = {9223--9232},
  publisher    = {{CVPR}},
  year         = {2023}
}

@inproceedings{wei2023surroundocc,
  author       = {Yi Wei and
                  Linqing Zhao and
                  Wenzhao Zheng and
                  Zheng Zhu and
                  Jie Zhou and
                  Jiwen Lu
                 },
  title        = {SurroundOcc: Multi-Camera 3D Occupancy Prediction for Autonomous Driving},
  booktitle    = {Proceedings of the {IEEE/CVF} International Conference on Computer Vision},
  pages        = {21672--21683},
  publisher    = {{ICCV}},
  year         = {2023}
}

@inproceedings{tian2023occ3d,
  author       = {Xiaoyu Tian and
                  Tao Jiang and
                  Longfei Yun and
                  Yucheng Mao and
                  Huitong Yang and
                  Yue Wang and
                  Yilun Wang and
                  Hang Zhao
                 },
  title        = {Occ3D: {A} Large-Scale 3D Occupancy Prediction Benchmark for Autonomous Driving},
  booktitle    = {Advances in Neural Information Processing Systems},
  volume       = {36},
  pages        = {64318--64330},
  publisher    = {{NeurIPS}},
  year         = {2023}
}

@misc{li2023fb,
  title   = {FB-OCC: 3D Occupancy Prediction based on Forward-Backward View Transformation},
  author  = {Zhiqi Li and Zhiding Yu and David Austin and Mingsheng Fang and Shiyi Lan and Jan Kautz and Jose M. Alvarez},
  eprint={2307.01492},
  archivePrefix="arxiv",
  year    = {2023}
}

@inproceedings{wang2024panoocc,
  author       = {Yuqi Wang and
                  Yuntao Chen and
                  Xingyu Liao and
                  Lue Fan and
                  Zhaoxiang Zhang
                 },
  title        = {PanoOcc: Unified Occupancy Representation for Camera-based 3D Panoptic Segmentation},
  booktitle    = {Proceedings of the {IEEE/CVF} Conference on Computer Vision and Pattern Recognition},
  pages        = {17158--17168},
  publisher    = {{CVPR}},
  year         = {2024}
}

@inproceedings{zhang2023occformer, 
  author={Zhang, Yunpeng and Zhu, Zheng and Du, Dalong},
  year={2023}, 
  title={Occformer: Dual-path transformer for vision-based 3d semantic occupancy prediction},
  booktitle={Proceedings of the {IEEE/CVF} International Conference on Computer Vision},
  pages={9433--9443}, 
  publisher={{ICCV}}, 
}

@misc{yu2023flashocc,
  title={Flashocc: Fast and memory-efficient occupancy prediction via channel-to-height plugin},
  author={Yu, Zichen and Shu, Changyong and Deng, Jiajun and Lu, Kangjie and Liu, Zongdai and Yu, Jiangyong and Yang, Dawei and Li, Hui and Chen, Yan}, 
  eprint={2311.12058},
  archivePrefix="arxiv", 
  year={2023}
}

@inproceedings{hou2024fastocc,
  author       = {Jiawei Hou and
                  Xiaoyan Li and
                  Wenhao Guan and
                  Gang Zhang and
                  Di Feng and
                  Yuheng Du and
                  Xiangyang Xue and
                  Jian Pu
                 },
  title        = {FastOcc: Accelerating 3D Occupancy Prediction by Fusing the 2D Bird's-Eye View and Perspective View},
  booktitle    = {{IEEE} International Conference on Robotics and Automation}, 
  pages        = {16425--16431}, 
  publisher    = {{ICRA}},
  year         = {2024}
}

@inproceedings{tang2024sparseocc,
  author       = {Pin Tang and
                  Zhongdao Wang and
                  Guoqing Wang and
                  Jilai Zheng and
                  Xiangxuan Ren and
                  Bailan Feng and
                  Chao Ma
                 },
  title        = {SparseOcc: Rethinking Sparse Latent Representation for Vision-Based Semantic Occupancy Prediction},
  booktitle    = {Proceedings of the {IEEE/CVF} Conference on Computer Vision and Pattern Recognition},
  pages        = {15035--15044},
  publisher    = {{CVPR}},
  year         = {2024}
}

@inproceedings{huang2024gaussianformer,
  author       = {Yuanhui Huang and
                  Wenzhao Zheng and
                  Yunpeng Zhang and
                  Jie Zhou and
                  Jiwen Lu
                 }, 
  title        = {GaussianFormer: Scene as Gaussians for Vision-Based 3D Semantic Occupancy Prediction},
  booktitle    = {Proceedings of the European Conference on Computer Vision},
  volume       = {15085},
  pages        = {376--393},
  publisher    = {{ECCV}},
  year         = {2024}
}

@inproceedings{huang2025gaussianformer2,
  author       = {Yuanhui Huang and
                  Amonnut Thammatadatrakoon and
                  Wenzhao Zheng and
                  Yunpeng Zhang and
                  Dalong Du and
                  Jiwen Lu
                 },
  title        = {GaussianFormer-2: Probabilistic Gaussian Superposition for Efficient 3D Occupancy Prediction},
  booktitle    = {Proceedings of the {IEEE/CVF} Conference on Computer Vision and Pattern Recognition},
  pages        = {27477--27486},
  publisher    = {CVPR},
  year         = {2025}
}

@misc{zhao2025gaussianformer3d,
  title={GaussianFormer3D: Multi-Modal Gaussian-based Semantic Occupancy Prediction with 3D Deformable Attention},
  author={Zhao, Lingjun and Wei, Sizhe and Hays, James and Gan, Lu},
  eprint={2505.10685},
  archivePrefix="arxiv",
  year={2025}
}

@misc{yan2025stgs,
  author       = {Xiaoyang Yan and
                  Muleilan Pei and
                  Shaojie Shen
                 },
  title        = {{ST-GS:} Vision-Based 3D Semantic Occupancy Prediction with Spatial-Temporal
                  Gaussian Splatting},
  eprint       = {2505.18992},
  archivePrefix= "arxiv",
  year         = {2025}

}

@inproceedings{leng2025stocc,
  title={Occupancy learning with spatiotemporal memory},
  author={Leng, Ziyang and Yang, Jiawei and Yi, Wenlong and Zhou, Bolei},
  booktitle={Proceedings of the IEEE/CVF International Conference on Computer Vision},
  pages={26569--26578}, 
  year={2025}
}

@misc{shi2025oneocc,
  title={OneOcc: Semantic Occupancy Prediction for Legged Robots with a Single Panoramic Camera},
  author={Shi, Hao and Wang, Ze and Guo, Shangwei and Duan, Mengfei and Wang, Song and Chen, Teng and Yang, Kailun and Wang, Lin and Wang, Kaiwei},
  eprint       = {2511.03571},
  archivePrefix= "arxiv",
  year         = {2025}
}

@inproceedings{li2022bevformer,
  author       = {Zhiqi Li and
                  Wenhai Wang and
                  Hongyang Li and
                  Enze Xie and
                  Chonghao Sima and
                  Tong Lu and
                  Yu Qiao and
                  Jifeng Dai},
  editor       = {Shai Avidan and
                  Gabriel J. Brostow and
                  Moustapha Ciss{\'{e}} and
                  Giovanni Maria Farinella and
                  Tal Hassner},
  title        = {BEVFormer: Learning Bird's-Eye-View Representation from Multi-camera
                  Images via Spatiotemporal Transformers},
  booktitle    = {Proceedings of the European Conference on Computer Vision},
  series       = {Lecture Notes in Computer Science},
  volume       = {13669},
  pages        = {1--18},
  publisher    = {{ECCV}},
  year         = {2022},
}

@misc{hang2021bevdet,
  author       = {Junjie Huang and
                  Guan Huang and
                  Zheng Zhu and
                  Dalong Du},
  title        = {BEVDet: High-performance Multi-camera 3D Object Detection in Bird-Eye-View},
  eprint       = {2112.11790},
  archivePrefix= "arxiv",
  year         = {2021}
}

@inproceedings{liu2023sparsebev,
  author       = {Haisong Liu and
                  Yao Teng and
                  Tao Lu and
                  Haiguang Wang and
                  Limin Wang
                 },
  title        = {SparseBEV: High-Performance Sparse 3D Object Detection from Multi-Camera
                  Videos},
  booktitle    = {Proceedings of the {IEEE/CVF} International Conference on Computer Vision},
  pages        = {18534--18544},
  publisher    = {{ICCV}},
  year         = {2023},
}

@inproceedings{song2017sscnet,
  author       = {Shuran Song and
                  Fisher Yu and
                  Andy Zeng and
                  Angel X. Chang and
                  Manolis Savva and
                  Thomas A. Funkhouser
                 },
  title        = {Semantic Scene Completion from a Single Depth Image}, 
  booktitle    = {Proceedings of the {IEEE} Conference on Computer Vision and Pattern Recognition}, 
  pages        = {190--198}, 
  publisher    = {CVPR}, 
  year         = {2017}, 
}

@inproceedings{wang2019forknet,
  author       = {Yida Wang and 
                  David Joseph Tan and 
                  Nassir Navab and 
                  Federico Tombari 
                 }, 
  title        = {ForkNet: Multi-Branch Volumetric Semantic Completion From a Single Depth Image},
  booktitle    = {Proceedings of the {IEEE/CVF} International Conference on Computer Vision},
  pages        = {8607--8616},
  publisher    = {{ICCV}},
  year         = {2019}
}

@inproceedings{wang2023semantic,
  author       = {Fengyun Wang and
                  Dong Zhang and
                  Hanwang Zhang and
                  Jinhui Tang and
                  Qianru Sun
                 },
  title        = {Semantic Scene Completion with Cleaner Self},
  booktitle    = {Proceedings of the {IEEE/CVF} Conference on Computer Vision and Pattern Recognition},
  pages        = {867--877},
  publisher    = {{CVPR}},
  year         = {2023}
}

@inproceedings{roldao2020lmscnet,
  author       = {Luis Rold{\~{a}}o and
                  Raoul de Charette and
                  Anne Verroust{-}Blondet
                 },
  title        = {LMSCNet: Lightweight Multiscale 3D Semantic Completion},
  booktitle    = {8th International Conference on 3D Vision},
  pages        = {111--119},
  publisher    = {{3DV}}, 
  year         = {2020}
}

@inproceedings{cao2024monoscene,
  author       = {Anh{-}Quan Cao and
                  Raoul de Charette
                 },
  title        = {MonoScene: Monocular 3D Semantic Scene Completion},
  booktitle    = {Proceedings of the {IEEE/CVF} Conference on Computer Vision and Pattern Recognition},
  pages        = {3981--3991},
  publisher    = {{CVPR}},
  year         = {2022}
}

@inproceedings{li2023voxformer,
  author       = {Yiming Li and
                  Zhiding Yu and
                  Christopher B. Choy and
                  Chaowei Xiao and
                  Jos{\'{e}} M. {\'{A}}lvarez and
                  Sanja Fidler and
                  Chen Feng and
                  Anima Anandkumar
                 },
  title        = {VoxFormer: Sparse Voxel Transformer for Camera-Based 3D Semantic Scene
                  Completion},
  booktitle    = {Proceedings of the {IEEE/CVF} Conference on Computer Vision and Pattern Recognition},
  pages        = {9087--9098},
  publisher    = {{CVPR}},
  year         = {2023}
}

@article{mei2024camera,
  author       = {Jianbiao Mei and
                  Yu Yang and
                  Mengmeng Wang and
                  Junyu Zhu and
                  Jongwon Ra and
                  Yukai Ma and
                  Laijian Li and
                  Yong Liu
                 },
  title        = {Camera-Based 3D Semantic Scene Completion With Sparse Guidance Network},
  journal      = {{IEEE} Transactions on Image Processing},
  volume       = {33},
  pages        = {5468--5481},
  year         = {2024}
}

@inproceedings{zhu2024CGFormer,
  author = {Yu, Zhu and Zhang, Runmin and Ying, Jiacheng and Yu, Junchen and Hu, Xiaohai and Luo, Lun and Cao, Si-Yuan and Shen, Hui-liang},
  title = {Context and Geometry Aware Voxel Transformer for Semantic Scene Completion},
  booktitle = {Advances in Neural Information Processing Systems},
  volume = {37},
  pages = {1531--1555},
  publisher={{NeurIPS}},
  year = {2024}
}

@inproceedings{jiang2024symphonize,
  author       = {Haoyi Jiang and
                  Tianheng Cheng and
                  Naiyu Gao and
                  Haoyang Zhang and
                  Tianwei Lin and
                  Wenyu Liu and
                  Xinggang Wang
                 },
  title        = {Symphonize 3D Semantic Scene Completion with Contextual Instance Queries},
  booktitle    = {Proceedings of the {IEEE/CVF} Conference on Computer Vision and Pattern Recognition},
  pages        = {20258--20267},
  publisher    = {{CVPR}},
  year         = {2024}
}

@misc{ruiqian2025splatssc,
  author        = {Rui Qian and
                   Haozhi Cao and
                   Tianchen Deng and
                   Shenghai Yuan and
                   Lihua Xie
                  },
  title         = {SplatSSC: Decoupled Depth-Guided Gaussian Splatting for Semantic Scene
                  Completion},
  eprint        = {2508.02261},
  archivePrefix = "arxiv",
  year          = {2025}, 
}

@inproceedings{cao2024pasco,
  author       = {Anh{-}Quan Cao and
                  Angela Dai and
                  Raoul de Charette
                 },
  title        = {PaSCo: Urban 3D Panoptic Scene Completion with Uncertainty Awareness},
  booktitle    = {Proceedings of the {IEEE/CVF} Conference on Computer Vision and Pattern Recognition},
  pages        = {14554--14564},
  publisher    = {{CVPR}}, 
  year         = {2024}, 
}

@inproceedings{severin2025occuq,
  author       = {Severin Heidrich and
                  Till Beemelmanns and
                  Alexey Nekrasov and
                  Bastian Leibe and
                  Lutz Eckstein},
  title        = {{OCCUQ:} Exploring Efficient Uncertainty Quantification for 3D Occupancy
                  Prediction},
  booktitle    = {{IEEE} International Conference on Robotics and Automation},
  pages        = {1--8},
  publisher    = {{ICRA}},
  year         = {2025}
}

@inproceedings{wu2024embodiedocc,
  author       = {Yuqi Wu and
                  Wenzhao Zheng and
                  Sicheng Zuo and
                  Yuanhui Huang and
                  Jie Zhou and
                  Jiwen Lu
                 },
  title        = {EmbodiedOcc: Embodied 3D Occupancy Prediction for Vision-based Online Scene Understanding},
  booktitle    = {Proceedings of the {IEEE/CVF} International Conference on Computer Vision},
  publisher    = {{ICCV}}, 
  year         = {2025},
}

@inproceedings{wang2025embodiedoccplusplus,
  title= {EmbodiedOcc\texttt{++}: Boosting Embodied 3D Occupancy Prediction with Plane Regularization and Uncertainty Sampler},
  author= {Wang, Hao and Wei, Xiaobao and Zhang, Xiaoan and Li, Jianing and Bai, Chengyu and Li, Ying and Lu, Ming and Zheng, Wenzhao and Zhang, Shanghang},
  booktitle= {Proceedings of the 33rd ACM International Conference on Multimedia},
  publisher= {{MM}}, 
  year= {2025},  
}

@misc{zhang2025roboocc,
  title={Roboocc: Enhancing the geometric and semantic scene understanding for robots},
  author={Zhang, Zhang and Zhang, Qiang and Cui, Wei and Shi, Shuai and Guo, Yijie and Han, Gang and Zhao, Wen and Ren, Hengle and Xu, Renjing and Tang, Jian},
  eprint={2504.14604},
  archivePrefix="arxiv",
  year={2025}
}

@article{xian2025discene,
  author       = {Xiang Li and
                  Yupeng Zheng and
                  Pengfei Li and
                  Yilun Chen and
                  Ya{-}Qin Zhang and
                  Wenchao Ding
                 },
  title        = {Enhancing Indoor Occupancy Prediction via Sparse Query-Based Multi-Level
                  Consistent Knowledge Distillation},
  journal      = {{IEEE} Robotics Autom. Lett.},
  volume       = {10},
  number       = {11},
  pages        = {11690--11697},
  year         = {2025},
}

@inproceedings{taolu2024scaffoldgs,
  author       = {Tao Lu and
                  Mulin Yu and
                  Linning Xu and
                  Yuanbo Xiangli and
                  Limin Wang and
                  Dahua Lin and
                  Bo Dai
                 },
  title        = {Scaffold-GS: Structured 3D Gaussians for View-Adaptive Rendering},
  booktitle    = {Proceedings of the {IEEE/CVF} Conference on Computer Vision and Pattern Recognition},
  pages        = {20654--20664},
  publisher    = {{CVPR}},
  year         = {2024},
}

@misc{jiang2025anysplat,
  title={AnySplat: Feed-forward 3D Gaussian Splatting from Unconstrained Views},
  author={Jiang, Lihan and Mao, Yucheng and Xu, Linning and Lu, Tao and Ren, Kerui and Jin, Yichen and Xu, Xudong and Yu, Mulin and Pang, Jiangmiao and Zhao, Feng and others},
  eprint={2505.23716},
  archivePrefix="arxiv",
  year={2025} 
}

@inproceedings{tan2019efficientnet,
  author       = {Mingxing Tan and
                  Quoc V. Le
                 }, 
  title        = {EfficientNet: Rethinking Model Scaling for Convolutional Neural Networks},
  booktitle    = {Proceedings of the 36th International Conference on Machine Learning},
  volume       = {97},
  pages        = {6105--6114},
  publisher    = {{PMLR}}, 
  year         = {2019}
}

@inproceedings{lin2017feature,
  author       = {Tsung{-}Yi Lin and
                  Piotr Doll{\'{a}}r and
                  Ross B. Girshick and
                  Kaiming He and
                  Bharath Hariharan and
                  Serge J. Belongie
                 }, 
  title        = {Feature Pyramid Networks for Object Detection},
  booktitle    = {Proceedings of the {IEEE} Conference on Computer Vision and Pattern Recognition},
  pages        = {936--944},
  publisher    = {{CVPR}},
  year         = {2017}
}

@inproceedings{yang2024depth,
  author={Yang, Lihe and Kang, Bingyi and Huang, Zilong and Zhao, Zhen and Xu, Xiaogang and Feng, Jiashi and Zhao, Hengshuang},
  title={Depth anything v2},
  booktitle={Advances in Neural Information Processing Systems},
  volume={37},
  pages={21875--21911},
  publisher={{NeurIPS}},
  year={2024}, 
}

@inproceedings{yu2024monocular,
  author={Yu, Hongxiao and Wang, Yuqi and Chen, Yuntao and Zhang, Zhaoxiang},
  title={Monocular occupancy prediction for scalable indoor scenes},
  booktitle={Proceedings of the European Conference on Computer Vision},
  volume= {15088}, 
  pages={38--54},
  publisher={ECCV}, 
  year={2024}, 
}

@inproceedings{ashish2017transformer,
  author       = {Ashish Vaswani and
                  Noam Shazeer and
                  Niki Parmar and
                  Jakob Uszkoreit and
                  Llion Jones and
                  Aidan N. Gomez and
                  Lukasz Kaiser and
                  Illia Polosukhin
                 },
  title        = {Attention is All you Need},
  booktitle    = {Advances in Neural Information Processing Systems},
  volume    = {30},
  pages        = {5998--6008},
  publisher    = {{NeurIPS}},
  year         = {2017}
}

@inproceedings{loshchilov2017decoupled,
  author       = {Ilya Loshchilov and
                  Frank Hutter
                 },
  title        = {Decoupled Weight Decay Regularization},
  booktitle    = {Proceedings of the Seventh International Conference on Learning Representations},
  publisher    = {{ICLR}},
  year         = {2019}
}

@misc{dao2023flashattention2,
  title={Flashattention-2: Faster attention with better parallelism and work partitioning},
  author={Dao, Tri},
  eprint={2307.08691},
  archivePrefix="arxiv",
  year={2023}
}

@misc{qiu2025gatedattn,
  title={Gated Attention for Large Language Models: Non-linearity, Sparsity, and Attention-Sink-Free},
  author={Qiu, Zihan and Wang, Zekun and Zheng, Bo and Huang, Zeyu and Wen, Kaiyue and Yang, Songlin and Men, Rui and Yu, Le and Huang, Fei and Huang, Suozhi and others},
  eprint={2505.06708},
  archivePrefix="arxiv",
  year={2025} 
}

@inproceedings{wang2024sfpnet,
  title={Sfpnet: Sparse focal point network for semantic segmentation on general lidar point clouds},
  author={Wang, Yanbo and Zhao, Wentao and Cao, Chuan and Deng, Tianchen and Wang, Jingchuan and Chen, Weidong},
  booktitle={European Conference on Computer Vision},
  pages={403--421}, 
  year={2024}, 
  publisher={{ECCV}}, 
}

@misc{deng2025mcnslammultiagentcollaborativeneural,
  title={MCN-SLAM: Multi-Agent Collaborative Neural SLAM with Hybrid Implicit Neural Scene Representation}, 
  author={Tianchen Deng and Guole Shen and Xun Chen and Shenghai Yuan and Hongming Shen and Guohao Peng and Zhenyu Wu and Jingchuan Wang and Lihua Xie and Danwei Wang and Hesheng Wang and Weidong Chen},
  eprint={2506.18678}, 
  archivePrefix="arxiv", 
  year={2025} 
}

@inproceedings{deng2025mne,
  title={Mne-slam: Multi-agent neural slam for mobile robots},
  author={Deng, Tianchen and Shen, Guole and Xun, Chen and Yuan, Shenghai and Jin, Tongxin and Shen, Hongming and Wang, Yanbo and Wang, Jingchuan and Wang, Hesheng and Wang, Danwei and others},
  booktitle={Proceedings of the Computer Vision and Pattern Recognition Conference},
  pages={1485--1494},
  year={2025},
  publisher={{CVPR}}, 
}

@InProceedings{deng2024plgslam,
    author    = {Deng, Tianchen and Shen, Guole and Qin, Tong and Wang, Jianyu and Zhao, Wentao and Wang, Jingchuan and Wang, Danwei and Chen, Weidong},
    title     = {PLGSLAM: Progressive Neural Scene Represenation with Local to Global Bundle Adjustment},
    booktitle = {Proceedings of the IEEE/CVF Conference on Computer Vision and Pattern Recognition (CVPR)},
    month     = {June},
    year      = {2024},
    pages     = {19657-19666}
}

@misc{deng2024compact,
  title={Compact 3D Gaussian Splatting For Dense Visual SLAM},
  author={Deng, Tianchen and Chen, Yaohui and Zhang, Leyan and Yang, Jianfei and Yuan, Shenghai and Wang, Danwei and Chen, Weidong},
  eprint={2403.11247}, 
  archivePrefix="arxiv", 
  year={2024} 
}

@article{deng2023long,
  title={Long-term visual simultaneous localization and mapping: Using a bayesian persistence filter-based global map prediction},
  author={Deng, Tianchen and Xie, Hongle and Wang, Jingchuan and Chen, Weidong},
  journal={IEEE Robotics \& Automation Magazine},
  volume={30},
  number={1},
  pages={36--49},
  year={2023},
  publisher={IEEE}
}

@misc{deng2025vpgs,
  title={VPGS-SLAM: Voxel-based Progressive 3D Gaussian SLAM in Large-Scale Scenes},
  author={Deng, Tianchen and Wu, Wenhua and He, Junjie and Pan, Yue and Jiang, Xirui and Yuan, Shenghai and Wang, Danwei and Wang, Hesheng and Chen, Weidong},
  eprint={2505.18992}, 
  archivePrefix="arxiv", 
  year={2025} 
}

@inproceedings{li2017focalloss,
  author       = {Tsung{-}Yi Lin and
                  Priya Goyal and
                  Ross B. Girshick and
                  Kaiming He and
                  Piotr Doll{\'{a}}r
                 }, 
  title        = {Focal Loss for Dense Object Detection},
  booktitle    = {Proceedings of the {IEEE} International Conference on Computer Vision},
  pages        = {2999--3007},
  publisher    = {{ICCV}},
  year         = {2017} 
}

@inproceedings{maxim2018lovaszloss,
  author       = {Maxim Berman and
                  Amal Rannen Triki and
                  Matthew B. Blaschko
                 },
  title        = {The Lov{\'{a}}sz-Softmax Loss: {A} Tractable Surrogate for the
                  Optimization of the Intersection-Over-Union Measure in Neural Networks},
  booktitle    = {Proceedings of the {IEEE} Conference on Computer Vision and Pattern Recognition},
  pages        = {4413--4421},
  publisher    = {{CVPR}},
  year         = {2018} 
}

@inproceedings{wang2024dust3r,
  title={Dust3r: Geometric 3d vision made easy},
  author={Wang, Shuzhe and Leroy, Vincent and Cabon, Yohann and Chidlovskii, Boris and Revaud, Jerome},
  booktitle={Proceedings of the IEEE/CVF Conference on Computer Vision and Pattern Recognition},
  pages={20697--20709},
  publisher={{CVPR}}, 
  year={2024}
}

@inproceedings{wang2025vggt,
  author       = {Jianyuan Wang and
                  Minghao Chen and
                  Nikita Karaev and
                  Andrea Vedaldi and
                  Christian Rupprecht and
                  David Novotn{\'{y}}
                 },
  title        = {{VGGT:} Visual Geometry Grounded Transformer},
  booktitle    = {Proceedings of the {IEEE/CVF} Conference on Computer Vision and Pattern Recognition},
  pages        = {5294--5306},
  publisher    = {{CVPR}},
  year         = {2025}
}

@InProceedings{wang2025cut3r,
    author    = {Wang, Qianqian and Zhang, Yifei and Holynski, Aleksander and Efros, Alexei A. and Kanazawa, Angjoo},
    title     = {Continuous 3D Perception Model with Persistent State},
    booktitle = {Proceedings of the IEEE/CVF Conference on Computer Vision and Pattern Recognition (CVPR)},
    pages     = {10510-10522}, 
    publisher = {{CVPR}}, 
    year      = {2025}
}

@misc{chen2025ttt3r,
  title={TTT3r: 3d reconstruction as test-time training},
  author={Chen, Xingyu and Chen, Yue and Xiu, Yuliang and Geiger, Andreas and Chen, Anpei},
  eprint={2509.26645},
  archivePrefix="arxiv",
  year={2025}
}

@inproceedings{gal2016dropout,
  title={Dropout as a bayesian approximation: Representing model uncertainty in deep learning},
  author={Gal, Yarin and Ghahramani, Zoubin},
  booktitle={international conference on machine learning},
  pages={1050--1059},
  publisher={PMLR}, 
  year={2016},
}

@inproceedings{lakshminarayanan2017simple,
  title={Simple and scalable predictive uncertainty estimation using deep ensembles},
  author={Lakshminarayanan, Balaji and Pritzel, Alexander and Blundell, Charles},
  booktitle={Advances in neural information processing systems},
  volume={30},  
  publisher={{NeurIPS}}, 
  year={2017} 
}

@inproceedings{havasi2021training,
  author       = {Marton Havasi and
                  Rodolphe Jenatton and
                  Stanislav Fort and
                  Jeremiah Zhe Liu and
                  Jasper Snoek and
                  Balaji Lakshminarayanan and
                  Andrew Mingbo Dai and
                  Dustin Tran
                 },
  title        = {Training independent subnetworks for robust prediction},
  booktitle    = {Proceedings of the Ninth International Conference on Learning Representations}, 
  publisher    = {{ICLR}},
  year         = {2021} 
}
}

\end{document}